\documentclass[10pt]{article}
\usepackage[utf8]{inputenc}
\usepackage{amsthm}
\usepackage{natbib}
\usepackage[left=2.5cm, right=2.5cm, top=2.25cm, bottom=2.25cm, headheight=20pt]{geometry}

\theoremstyle{definition}
\newtheorem{theorem}{Theorem}

\usepackage{amsmath,amssymb}
\usepackage{hyperref} 
\usepackage{lscape}
\usepackage{arydshln}
\usepackage{booktabs}
\usepackage{verbatim}
\usepackage{enumitem}
\usepackage{bm}
\usepackage{authblk}
\DeclareMathAlphabet{\mathpzc}{OT1}{pzc}{m}{it}
\usepackage{wasysym}
\usepackage{esint}
\usepackage{xcolor}
\usepackage{tikz}
\usetikzlibrary{shapes,backgrounds,arrows,chains,matrix,positioning,scopes}
\usepackage{caption}
\usepackage{titlesec}

\usepackage{lineno}

\newcommand{\R}{\mathbb{R}}

\newcommand{\xvec}{\mbox{\bf x}}

\newcommand{\Xvec}{\mbox{\bf X}}
\newcommand{\Yvec}{\mbox{\bf Y}}
\newcommand{\Zvec}{\mbox{\bf Z}}

\newcommand{\disp}{\displaystyle}
\newcommand{\sigmat}{\mbox{\boldmath $\Sigma$}}
\newcommand{\muvec}{\mbox{\boldmath $\mu$}}
\newcommand{\alphavec}{\mbox{\boldmath $\alpha$}}

\title{On high-dimensional modifications of the nearest neighbor classifier}
\author{Annesha Ghosh$^1$, Deep Ghoshal$^2$, Bilol Banerjee$^1$, and Anil K. Ghosh$^1$} 
\affil{$^1$ Theoretical Statistics and Mathematics Unit,
Indian Statistical Institute, India\\
$^2$ Department of Statistics, University of Illinois at Urbana-Champaign, USA. \\
E-mail*: anneshaghosh659@gmail.com}

\begin{document}

\maketitle

\begin{abstract}
   
Nearest neighbor classifier is arguably the most simple and popular nonparametric classifier available in the literature. However, due to the concentration of pairwise distances and the violation of the neighborhood structure, this classifier often suffers in high-dimension, low-sample size (HDLSS) situations, especially when the scale difference between the competing classes dominates their location difference. Several attempts have been made in the literature to take care of this problem. In this article, we discuss some of {those} existing methods and propose some new ones. We carry out some theoretical investigations in this regard and analyze several simulated and benchmark datasets to compare the empirical performances of {our} proposed methods with some of the existing ones.

\end{abstract}

\section{Introduction}

In supervised classification, we use a training set of labeled observations from different competing classes to form a decision rule for classifying unlabeled test set observations as accurately as possible. Starting from \cite{fisher1936use}, \cite{rao1948utilization} and \cite{fix1951}, several parametric as well as nonparametric classifiers have been developed for this purpose \citep[see, e.g.,][]{duda2006pattern, hastie2009elements}. Among them,
the nearest neighbor classifier \citep[see, e.g.,][]{cover1967nearest} is perhaps the most popular one. The $k$-nearest neighbor classifier ($k$-NN) classifies an observation $\xvec$ to the class having the maximum number of representatives among the $k$ nearest neighbors of $\xvec$. 
This classifier works well if the training sample size is large compared to the dimension of the data. For a suitable choice of $k$ (which increases with the training sample size at an appropriate rate), under some mild regularity conditions, the misclassification rate of the $k$-NN classifier 
converges to the Bayes risk (i.e., the misclassification rate of the Bayes classifier) as the training sample size grows to infinity \citep[see, e.g.][]{devroye2013probabilistic, hall2008choice}. 
However, like other nonparametric methods, this classifier also suffers from the curse of dimensionality \citep[see, e.g.,][]{carrerira2009review}, especially when the dimension of the data is much larger than the training sample size. 
In such high-dimension, low-sample-size (HDLSS) situations, the concentration of pairwise distances \cite[see, e.g.,][]{hall2005geometric, franccois2007concentration}, presence of hubs and the violation of the neighborhood structure \cite[see, e.g.,][]{radovanovic2010hubs, pal2016high} often have adverse effects on the performance of the nearest neighbor classifier. 

To demonstrate this, we consider some simple examples involving two $d$-dimensional normal distributions. Descriptions of these examples are given below.

{\textbf{Examples 1 - 3:} {\it In these three examples, the first class has a normal distribution with the mean vector ${\bf 0}_d=(0,0,\ldots,0)^{\top}$ 
and the dispersion matrix ${\bf I}_d$ (the $d \times d$ identity matrix), while the second class has the mean vector $\mu {\bf 1}_d=\mu(1,1,\ldots,1)^{\top}=(\mu,\mu,\ldots,\mu)^{\top}$
and the dispersion matrix $\sigma^2{\bf I}_d$. In Example 1, we consider a location problem where we take $\mu=1$ and $\sigma=1$. Example 2 deals with a location-scale problem with $\mu=1$ and $\sigma=2$. As Example 3, we choose a scale problem, where $\mu$ and $\sigma$ are taken as $0$ and $2$, respectively.}

In each of these examples, we carry out our experiment for $7$ different choices of $d$  ranging between $10$ and $1000$ {($d$= $10, 20, 50, 100, 200, 500$ and $1000$)}. In each case, taking an equal number of observations from the two competing classes, we form the training and test sets of size 50 and 500, respectively.
This is done 100 times, and the average test set misclassification rates of the $1$-NN classifier
over these 100 trials are reported in Figure \ref{fig:1-3}.

Note that in each of these examples, the distribution of each measurement variable differs in two competing classes. So, each of them contains information about class separability, and as a result, the separability between the two classes increases with the dimension. One can check that in each of these examples, the Bayes risk converges to $0$ as the dimension grows. Therefore, the misclassification rate of any good classifier is also expected to go down as the dimension increases. We observed the same for the $1$-NN classifier in Example 1 (location problem), but surprisingly, in the other two cases, its misclassification rates were close to 0.5 in high dimensions. 

A careful investigation explains the reasons for this diametrically opposite behavior. Let
$\{\mathbf{X}_1,\mathbf{X}_2,.....,\mathbf{X}_{n1}\}$
and $\{\mathbf{Y}_1,\mathbf{Y}_2,.....,\mathbf{Y}_{n_2}\}$ be the training samples from two competing classes (here we have $n_1=n_2=25)$ $N({\bf 0}_d,{\bf I}_d)$ and $N(\mu {\bf 1}_d,\sigma^2{\bf I}_d)$, respectively. Now, for a test case $\mathbf{Z}$ from $N({\bf 0}_d,{\bf I}_d)$, one can show that
for each $i=1,2,\ldots,n_1$, $\frac{1}{d}\|\Zvec-\Xvec_i\|^2$, being the average of independent and identically distributed (i.i.d.) random variables, converges in probability to $2$ as $d$ increases to infinity. Similarly, it can be shown that for  each $i=1,2,\ldots,n_2$, $\frac{1}{d}\|\Zvec-\Yvec_i\|^2 \stackrel{P}{\rightarrow}{1+\mu^2+\sigma^2}$. So, $\Zvec$ is correctly classified by the $1$-NN classifier (or any $k$-NN classifier with $k\le \min\{n_1,n_2\}$) if 
$\mu^2+\sigma^2>1$, Note that it was the case in all
three examples. So, all observations from $N({\bf 0}_d,{\bf I}_d)$ were correctly classified. But for a test case ${\mathbf{Z}}^{\prime}$ 
from $N(\mu{\bf 1}_d,\sigma^2{\bf I}_d)$, we have $\frac{1}{d}\|\Zvec^{\prime}-\Xvec_i\|^2 \stackrel{P}{\rightarrow}{1+\mu^2+\sigma^2}$ for $i=1,2,\ldots,n_1$ and $\frac{1}{d}\|\Zvec^{\prime}-\Yvec_i\|^2 \stackrel{P}{\rightarrow}{2\sigma^2}$ for $i=1,2,\ldots,n_2$. So, it is correctly classified if and only if $\sigma^2<1+\mu^2$. This condition was satisfied in Example 1, but not in the other two cases. Because of this violation of the neighborhood structure (where observations from one class have all neighbors from other classes), in Examples 2 and 3, the $1$-NN classifier misclassified all observations from $N(\mu{\bf 1}_d,\sigma^2{\bf I}_d)$ and had misclassification rates close to $0.5$.

This phenomenon of distance concentration in high dimension was observed by \cite{hall2005geometric} for Euclidean distances and \cite{franccois2007concentration} for fractional distances. \cite{hall2005geometric} also studied the high dimensional behavior of some popular classifiers and observed this undesirable behavior of the nearest neighbor classifier. To take care of this problem, \cite{chan2009scale} proposed an adjustment for the scale difference between the competing classes. They suggested to compute
\begin{align*}
& \rho_1(\Zvec,\Xvec_i)=\|\Zvec - \Xvec_i\|^2- \frac{1}{2} {\binom{n_1}{2}^{-1}} \sum_{s < t}\|\Xvec_s-\Xvec_t\|^2  \mbox{ 
 for } i=1,2,\ldots,n_1,\\
&\rho_2(\Zvec,\Yvec_i)=\| \Zvec- \Yvec_i\|^2- \frac{1}{2} \binom{n_2}{2}^{-1} \sum_{s < t}\|\Yvec_s-\Yvec_t\|^2 \mbox{ 
 for } i=1,2,\ldots,n_2
\end{align*} 
and classify $\Zvec$ to the first (respectively, second) class if $\min \rho_1(\Zvec,\Xvec_i)<\min  \rho_2(\Zvec,Y_i)$ (respectively,
 $\min \rho_1(\Zvec,\Xvec_i)>\min  \rho_2(\Zvec,Y_i)$). Note that without the scale adjustments
 (second terms on the right-hand side of the equations), it turns out to be the usual $1$-NN classifier. Figure \ref{fig:1-3} also shows the performance of this classifier (we refer to it as the CH classifier) in Examples 1-3. In Example 1, it performed like the $1$-NN classifier. Interestingly, in Example 2, while the $1$-NN classifier failed, this scale adjustment led to improved performance by the CH classifier in high dimensions. But in Example 3, like the $1$-NN classifier, it also misclassified almost 50\% observations. Note that for any $\Zvec$ from $N({\bf 0}_d, {\bf I}_d)$, here $\rho_1(\Zvec,\Xvec_i)/d \stackrel{P}{\rightarrow}{1}$ $(i=1,2,\ldots,n_1)$ and $\rho_2(\Zvec,\Yvec_i)/d \stackrel{P}{\rightarrow}{1+\mu^2}$ $(i=1,2,\ldots,n_2)$ as $d$ increases. So, it is correctly classified if $\mu^2> 0$. Again, for any $\Zvec^{\prime}$ from $N(\mu{\bf 1}_d,\sigma^2{\bf I}_d)$, we have $\rho_1(\Zvec^{\prime},\Xvec_i)/d \stackrel{P}{\rightarrow}{\mu^2+\sigma^2}$ for $i=1,2,\ldots,n_1$ and $\rho_2(\Zvec^{\prime},\Yvec_i)/d \stackrel{P}{\rightarrow}{\sigma^2}$ for $i=1,2,\ldots,n_2$. So, here also, we need $\mu^2>0$ for correct classification. In Examples 1 and 2, we had $\mu^2>0$. So, the CH classifier performed well in those two examples for large values of $d$. But in Example 3, where we had $\mu^2=0$, it misclassified almost 50\% observations. 
This example shows that the CH classifier may fail to discriminate between two high-dimensional distributions differing only in their scales.
 
 \begin{figure}[!b]
\centering
\setlength{\tabcolsep}{0pt}
\begin{tabular}{ccc}
(a) Ex. 1: Location problem & (b) Ex. 2: Location-scale & (c) Ex. 3: Scale problem  \\
  ($\mu=1$, $\sigma=1$) &
~~~~problem ($\mu=1$, $\sigma=2$) &
~~~~  ($\mu=0$, $\sigma=2$) \\
\includegraphics[width=0.32\linewidth]{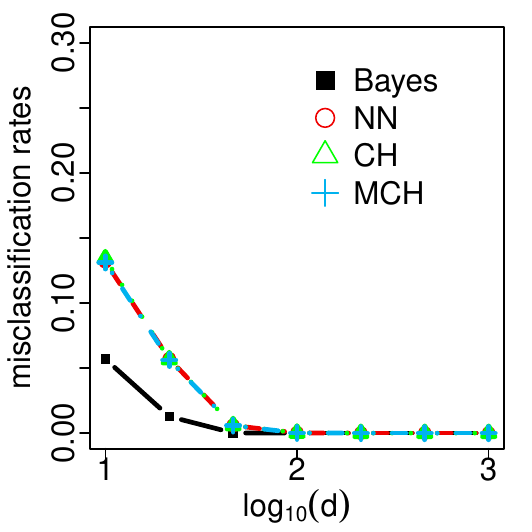} & \includegraphics[width=0.32\linewidth]{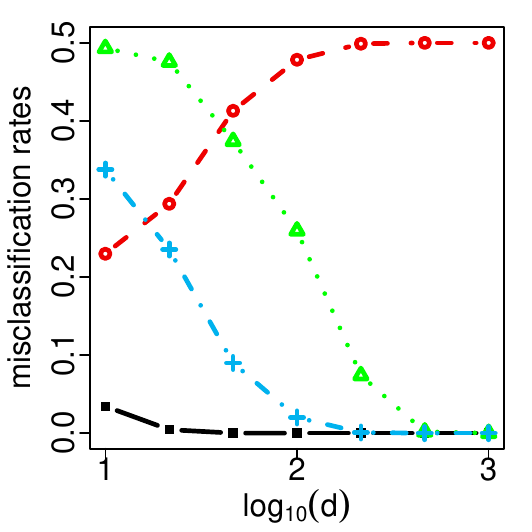} & \includegraphics[width=0.32\linewidth]{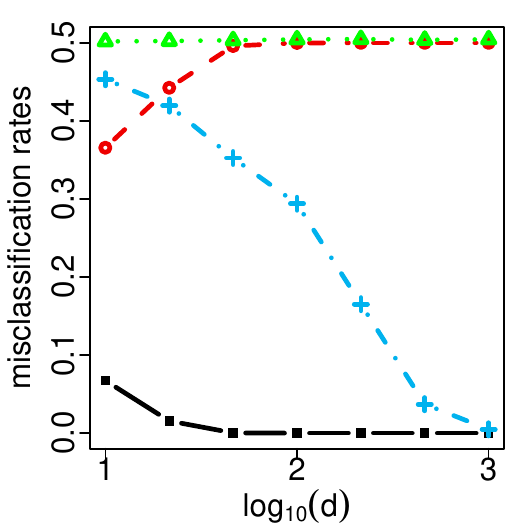} \\
\end{tabular}
\caption{Misclassification rates of Bayes, NN, CH and MCH classifiers in Examples 1-3.\label{fig:1-3}}
\end{figure}

 However, if we slightly modify \cite{chan2009scale}'s proposal of scale adjustment, we can take care of high dimensional scale problems as well. Our modified version (which we refer to as the Modified Chan and Hall classifier or the MCH classifier) had excellent performance in all three examples (see Figure \ref{fig:1-3}), especially for large values of $d$. In Section 2, we propose this modification and carry out some theoretical and numerical studies to understand the high-dimensional behavior of the resulting classifier.

An alternative strategy to deal with any high-dimensional problem is to reduce the dimension of the data and work on the reduced subspace. The simplest method of dimension reduction is to consider some random linear projections \citep[see, e.g.,][and the references therein]{fern2003random,fradkin2003experiments,vempala2005random}, and we can adopt that method for nearest neighbor classification as well. Another popular approach is to use projections based on principal component analysis \citep[see, e.g.,][]{deegalla2006reducing,macionczyk2023analyzing}. But as pointed out in \cite{dutta2016some}, these methods often lead to poor performance in high-dimensional classification problems. For instance, from our description, it is quite clear that neither the principle component directions nor the random projections are meaningful in Example 3. Also getting consistent estimates of the principal components in high dimension is challenging \citep[see, e.g.,][]{jung2009pca}. Other approaches towards nearest neighbor classification of high-dimensional data include those based on mean absolute difference of distances \citep[see,e.g.,][]{pal2016high,roy2022generalizations}, hubness-based fuzzy measures \citep[see, e.g.,][]{tomavsev2014hubness} and distance metrics learning \citep[see, e.g.][]{weinberger2009distance}. \cite{chan2009robust} proposed a robust version of the nearest neighbor method for classifying high-dimensional data, but their method can be used only for a specific type of two-class location problem.
 Instead of using random projections or principal components, \cite{dutta2016some} suggested extracting some distance-based features from the data and performing nearest neighbor classification based on those features. They proposed two such methods, one using transformation based on average distances (TRAD) and the other using transformation based on inter-point distances (TRIPD). We briefly discuss these two methods in Section 3 and also propose some other methods for selecting distance-based features for nearest neighbor calssification of high dimensional data. A comparative discussion of these methods is also given in this section based on our analysis of some simulated data sets. Some benchmark data sets are analyzed in Section 4
 to compare the performances of these methods with two popular state-of-the-art classifiers, support vector machines \citep[see, e.g.][]{christianini2003support,steinwart2008support,scholkopf2018learning} and random forest \citep[see, e.g.,][]{breiman2001random,genuer2020random}, which are known to perform well for high dimensional data. Finally, a brief summary of the work and some concluding remarks are given in Section 5. All proofs and mathematical details are given in the Appendix.

\section{Modified Scale-adjusted Nearest Neighbor Classifier}

We have seen that while the $1$-NN classifier failed in Examples 2 and 3, the scale-adjusted CH classifier worked well in Example 2 when the dimension was large. But, in Example 3, this scale adjustment could not improve the performance of the $1$-NN classifier. This motivates us to look for a modified scale adjustment. We define 
\begin{align*}
& \rho_1^{*}(\Zvec,\Xvec_i)=\|\Zvec - \Xvec_i\|- \frac{1}{2} {\binom{n_1}{2}^{-1}} \sum_{s < t}\|\Xvec_s-\Xvec_t\|  \mbox{ 
 for } i=1,2,\ldots,n_1,\\
&\rho_2^{*}(\Zvec,\Yvec_i)=\| \Zvec- \Yvec_i\|- \frac{1}{2} \binom{n_2}{2}^{-1} \sum_{s < t}\|\Yvec_s-\Yvec_t\| \mbox{ 
 for } i=1,2,\ldots,n_2
\end{align*} 
and classify a test set observation $\Zvec$ to the first (respectively, second) class if $\min \rho_1^{*}(\Zvec,\Xvec_i)$ is smaller (respectively, larger) than
$\min~\rho^{*}_2(\Zvec,\Yvec_i)$. Figure \ref{fig:1-3} shows that this modified scale adjusted nearest neighbor classifier (henceforth referred to as the Modified Chan and Hall classifier or the MCH classifier) had excellent performance in high dimensions in all three examples. A small theoretical analysis explains the reasons for its superior performance. 

Following our previous discussion on distance convergence, one can show that for a test set observation 
$\Zvec$ from $N({\bf 0}_d,{\bf I}_d)$, as $d$ tends to infinity, we have  $\rho_1^{\ast}(\Zvec,\Xvec_i)/\sqrt{d} \stackrel{P}{\rightarrow} 1/\sqrt{2}$ for $i=1,2,\ldots,n_1$, while 
$\rho_2^{\ast}(\Zvec,\Yvec_i)/\sqrt{d} \stackrel{P}{\rightarrow} \sqrt{1+\mu^2+\sigma^2}-\sigma/\sqrt{2}$ for $i=1,2,\ldots,n_2$. So, it is correctly classified if 
$\sqrt{1+\mu^2+\sigma^2}>(\sigma+1)/\sqrt{2} \Leftrightarrow 1+\mu^2+\sigma^2>(\sigma+1)^2/2 \Leftrightarrow \mu^2+\frac{1}{2}(\sigma-1)^2>0$. Again for an observation $\Zvec^{\prime}$
from $N(\mu{\bf 1}_d, \sigma^2{\bf I}_d)$, as $d \rightarrow \infty$, we have 
$\rho_1^{\ast}(\Zvec^{\prime},\Xvec_i)/\sqrt{d} \stackrel{P}{\rightarrow} \sqrt{1+\mu^2+\sigma^2}-1/\sqrt{2}$ for $i=1,2,\ldots,n_1$ and
$\rho_2^{\ast}(\Zvec^{\prime},\Yvec_i)/\sqrt{d} \stackrel{P}{\rightarrow} \sigma/\sqrt{2}$ for $i=1,2,\ldots,n_2$. So, here also, $\Zvec^{\prime}$ is correctly classified if $\mu^2+\frac{1}{2}(\sigma-1)^2>0$. This inequality holds in all three examples considered in Section 1. This was the reason for the excellent performance of the MCH classifier in high dimensions.

Like the usual nearest neighbor classifier, multi-class generalizations of CH and MCH classifiers are quite straightforward. If there are $J$ competing classes $F_1,F_2,\ldots,F_J$ with the training samples $\{\Xvec_{j1},\Xvec_{j2},\ldots,\Xvec_{jn_j}\}$ from the $j$-th class, ($j=1,2,\ldots,J$), for classifying a test case $\Zvec$ by the CH classifier, we can compute
$$\rho_j(\Zvec,\Xvec_{ji})=\|\Zvec - \Xvec_{ji}\|^2- \frac{1}{2} {\binom{n_j}{2}^{-1}} \sum_{s <t }\|\Xvec_{js}-\Xvec_{jt}\|^2  \mbox{ for } j=1,2,\ldots,J, ~i=1,2,\ldots,n_j$$ and assign  
 $\Zvec$ to class $j_0$ if $\min\limits_{1 \le i \le n_{j_0}} \rho_{j_0}(\Zvec,\Xvec_{j_0i}) < \min\limits_{1 \le i \le n_j} \rho_j(\Zvec,\Xvec_{ji})$ for all $j \neq j_0$.
Similarly, for the MCH classifier, one can compute
$$\rho^{\ast}_j(\Zvec,\Xvec_{ji})=\|\Zvec - \Xvec_{ji}\|- \frac{1}{2} {\binom{n_j}{2}^{-1}} \sum_{s <t }\|\Xvec_{js}-\Xvec_{jt}\|  \mbox{ for } j=1,2,\ldots,J, ~i=1,2,\ldots,n_j$$ and assign  
 $\Zvec$ to class $j_0$ if $\min\limits_{1 \le i \le n_{j_0}} \rho^{\ast}_{j_0}(\Zvec,\Xvec_{j_0i}) < \min\limits_{1 \le i \le n_j} \rho^{\ast}_j(\Zvec,\Xvec_{ji})$ for all $j \neq j_0$.
 
For the sake of simplicity, in Examples 1-3, we considered binary classification problems involving two normal distributions each having i.i.d. measurement variables.  
Now, one may be curious to know how CH and MCH classifiers perform in high-dimensional multi-class classification problems involving more general class distributions with possibly dependent and non-identically distributed measurement variables. 
For this investigation, we consider the following assumptions.

\begin{enumerate}
    \item[(A1)] In each of the $J$ competing classes, the measurement variables have uniformly bounded fourth moments.
    \item[(A2)] If $\Xvec = (X_1,\ldots,X_d)^{\top}\sim F_j$ and $\Yvec = (Y_1,\ldots,Y_d)^{\top}\sim F_i$ ($1\le j,i\le J$) are independent, for ${\bf U}=\Xvec-\Yvec$,  $\sum_{r \neq s}|Corr(U^2_r,U^2_s)|$
    is of the order $o(d^2)$.
    \item[(A3)] Let $\muvec_{j}$ and $\sigmat_{j}$ be the mean vector and the dispersion matrix of  $F_j$ ($1 \le j \le J$). For each $j=1,\ldots,J$, 
    there exists a constant $\sigma_j^2$ such that $\mbox{trace}(\sigmat_J)/d \rightarrow \sigma_j^2$ 
    as $d \rightarrow \infty$. Also, for each $i\neq j$, there exists a constant $\nu_{ji}^2$ such that $\|\muvec_j-\muvec_i\|^2/d \rightarrow \nu_{ji}^2$ as $d \rightarrow \infty$.
\end{enumerate}

Under (A1) and (A2), we have the weak law of large numbers (WLLN) \citep[see, e.g.,][]{feller1991introduction} for the sequence of
possibly dependent and non-identically distributed 
random variables $\{U^2_q:q\ge 1\}$, i.e., $\left|\frac{1}{d}\|{\bf U}\|^2-E\big(\frac{1}{d}\|{\bf U}\|^2\big)\right| \stackrel{P}{\rightarrow}0$ as $d \rightarrow \infty$ (note that if the measurement variables are i.i.d., as they were in Examples 1-3, the WLLN holds under the second moment assumption, (A1) and (A2) are not needed there). Assumption (A3) gives the limiting value of $E\big(\frac{1}{d}\|{\bf U}\|^2\big)$ and hence that of $\frac{1}{d}\|{\bf U}\|^2=\frac{1}{d}\|\Xvec-\Yvec\|^2$ for $\Xvec \sim F_j$ and $\Yvec \sim F_i$ ($1\le j,i \le J$). So, under (A1)-(A3), we have high-dimensional convergence of all pairwise distances and their limiting values (see Lemma 1 in Appendix). 
These assumptions are quite standard in the HDLSS literature. \cite{hall2005geometric} considered the $d$-dimensional observations as time series truncated at time $d$, and in addition to (A1) and (A3), they assumed the $\rho$-mixing property of the time series to study the high dimensional behavior of some popular classifier as $d$ increases. Note that (A2) holds under that $\rho$-mixing condition.  \cite{franccois2007concentration} observed that for high-dimensional data with highly correlated or dependent measurement variables, pairwise distances are less concentrated than if all variables are independent. They claimed that the distance concentration phenomenon depends
on the intrinsic dimension \citep[see, e.g.,][]{levina2004maximum,camastra2016intrinsic} of the data, instead of the dimension of the embedding space. So, in order to have distance concentration in high dimensions, one needs high intrinsic
dimensionality of the data or weak dependence among the measurement variables. The assumption (A2) ensures
that weak dependence. Some other similar relevant conditions for the convergence of pairwise distances can be found in \citep{ahn2007high,jung2009pca,sarkar2019perfect,yata2020geometric,banerjee2022high}.
Under (A1)-(A3), we have the following theorem on the misclassification rates of the usual nearest neighbor, CH and MCH classifiers. 

\begin{theorem}
    If $J$ competing classes satisfy assumptions (A1)-(A3), and there are at least two observations from each of them (i.e, $n_j\ge 2$ for all $j=1,2,\ldots, J$), then we have the following results.
    \begin{enumerate}
        \item[(a)] If $\nu_{ji}^2>|\sigma_j^2-\sigma_i^2|$ for all  $j \neq i$, the misclassification probability of the $k$-nearest neighbor classifier with $k<\min\{n_1,\ldots,n_J\}$ converges to $0$ as the dimension $d$ grows to infinity. However, if $\nu_{ji}^2<|\sigma_j^2-\sigma_i^2|$ for some  $j \neq i$, all observations from at least one class is misclassified with probability tending to $1$ as $d$ diverges to infinity.
        
        \item[(b)]  If   $\nu_{ji}^2>0$ for $j\neq i$, the misclassification probability of the CH classifier converges to $0$ as $d$ grows to infinity. 
        \item[(c)] Suppose that for all $j \neq i$, either   $\nu_{ji}^2>0$ or $\sigma_j^2\neq \sigma_i^2$. Then, the misclassification probability of the MCH classifier converges to $0$ as $d$ grows to infinity. 
    \end{enumerate}
\end{theorem}

Note that in Examples 1-3, we had $\nu_{12}^2=\mu^2$, $\sigma_1^2=1$ and $\sigma_2^2=\sigma^2$. The condition $\nu_{12}^2>|\sigma_1^2-\sigma_2^2|$ was violated in Examples 2 and 3, whereas the condition $\nu_{12}^2>0$ was also violated in Example 3. We had poor performance of NN and CH classifiers in these respective cases. But the condition $\nu_{12}^2>0$ or $\sigma_1^2 \neq \sigma_2^2$ was satisfied in all three examples. Consequently, the MCH classifier had good high-dimensional performance.

\begin{figure}[h]
\centering
\setlength{\tabcolsep}{1pt}
\begin{tabular}{ccc}
(a) Ex. 4: Two mixture & (b) Ex. 5: Uniform vs. & (c) Ex. 6: Normal$({\bf 0}_d,3{\bf I}_d)$ vs.  \\
 normal distributions & mixture of uniforms & standard $t$ with $3$ d.f.\\
\includegraphics[width=0.33\linewidth]{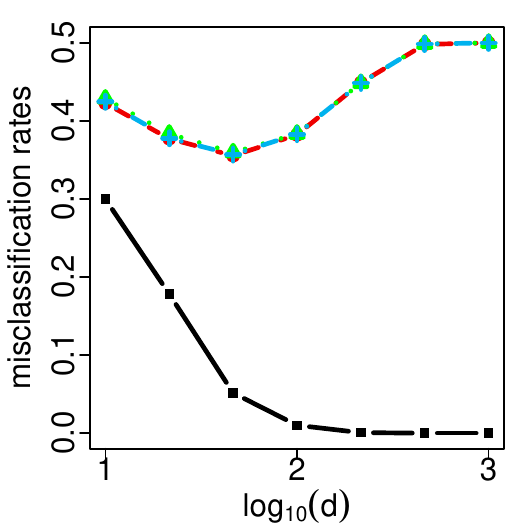} & \includegraphics[width=0.33\linewidth]{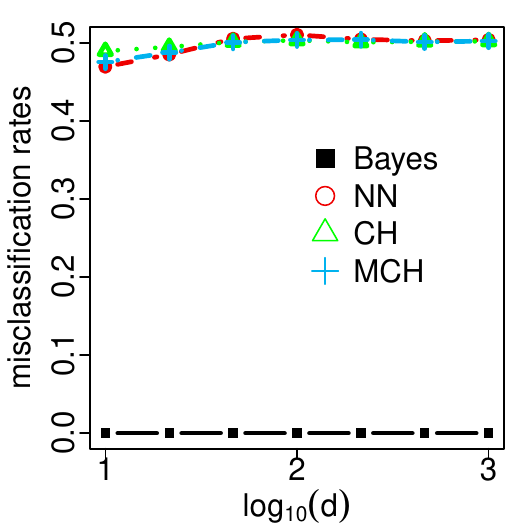} & \includegraphics[width=0.33\linewidth]{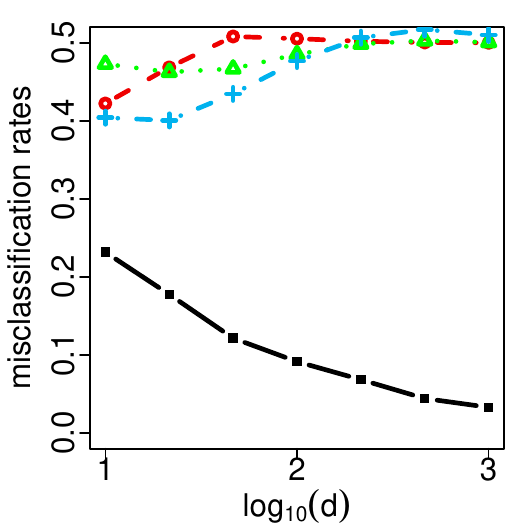} \\ 
\end{tabular}
\caption{Misclassification rates of Bayes, NN, CH, and MCH classifiers in Examples 4-6.\label{fig:4-6}}
\end{figure}

Now, we consider three more examples (Examples 4-6) to investigate how the MCH classifier performs when at least one of the assumptions of Theorem 1 is violated.

{\textbf{Example 4:} {\it Each of the two classes is an equal mixture of two normal distributions. 
While one class is a mixture of $N({\bf 0}_d, {\bf I}_d)$ and
$N({\bf 1}_d, 2{\bf I}_d)$, the other one is a mixture of 
$N(\alphavec_d, {\bf I}_d)$ and $N(({\bf 1}_d-\alphavec_d), 2{\bf I}_d)$, where $\alphavec_d$ is a $d$-dimensional vector with entries $0$ and $1$ at even and odd places, respectively.}


 {\textbf{Example 5:} {\it Here the two classes are $U_{d}(1, 1.5)$ and 
an equal mixture of $U_{d}(0.5, 1)$ and $U_{d}(1.5, 2)$, where $U_{d}(a, b)$ denotes the $d$-dimensional uniform
distribution over the region $\{\xvec \in {\mathbb R}^d: a \le \|S^{1/2}x\| \le b\}$ for $S=0.5{\bf I}_d+0.5{\bf 1}_d{\bf 1}_d^{\top}$.}

{Examples 4 and 5 are dealing with mixture distributions. Here, (A1)-(A3) hold for each of the four sub-classes, but (A2) is violated for both competing classes. 
We also consider the following example:}

{\textbf{Example 6:} {\it In this example, the two competing classes are $N({\bf 0}_d, 3{\bf I}_d)$ and  the standard multivariate $t$-distribution with $3$ degrees of freedom.}

In Example 6, (A2) is violated for the $t$-distribution. Moreover, since the two classes have the same mean vector and the same dispersion matrix, we have $\nu_{12}^2=0$ and $\sigma_1^2=\sigma_2^2$. For each example, we consider different values of $d$ ranging between $10$ and $1000$, and in each case, we form training and test samples of size 50 and 500, respectively, taking an equal number of observations from each class. Each experiment is repeated $100$ times to compute the average test set misclassification rates of different classifiers, and they are reported in Figure \ref{fig:4-6}. In these examples, NN, CH and MCH classifiers, all had poor performance, and they had misclassification rates close to $0.5$ in high dimensions.
These examples clearly show the necessity to develop some new methods for high dimensional nearest neighbor classification, particularly for the examples involving mixture distributions. In the next section, we propose and discuss some methods for this purpose.





\section{Nearest neighbor classification using distance-based features}

In the previous sections, we have seen that for high dimensional classification based on nearest neighbors, the scale adjustment methods (CH and MCH) may not always be helpful. To take care of this problem, 
we suggest extracting some distance-based features from the data and constructing a suitable classifier on that feature space.

\subsection{Classification based on minimum distances}

\begin{figure}[!b]
\centering
\setlength{\tabcolsep}{-2pt}
\begin{tabular}{ccc}
(a) NN and CH classifiers & (b) NN and MCH classifiers & (c) MDist classifier\\
\includegraphics[width=0.34\linewidth]{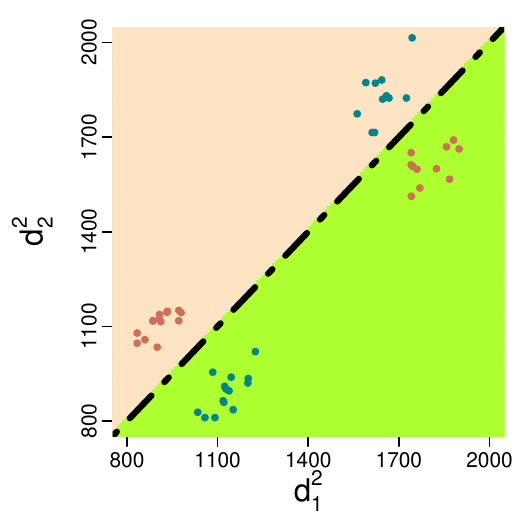} & \includegraphics[width=0.34\linewidth]{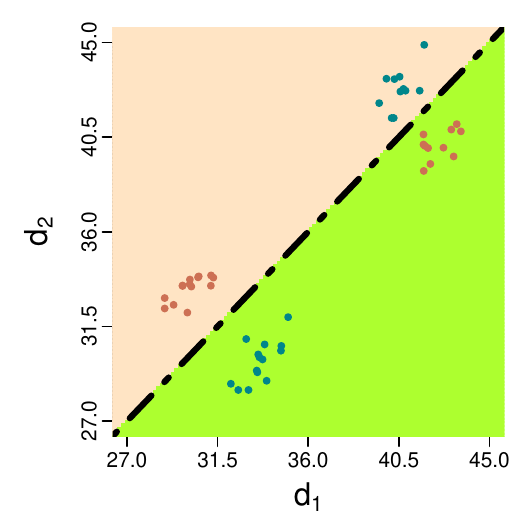} & \includegraphics[width=0.34\linewidth]{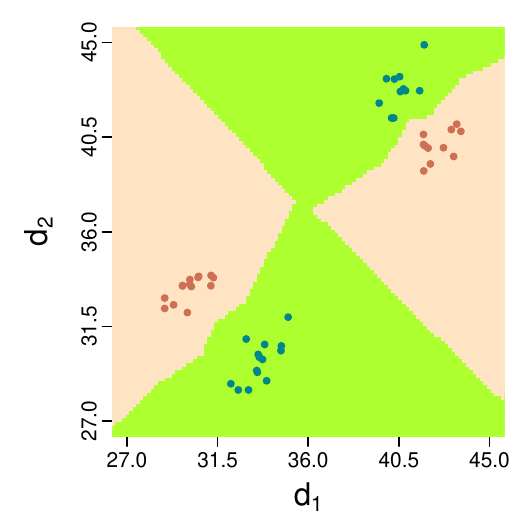}  \\ 
\includegraphics[width=0.34\linewidth]{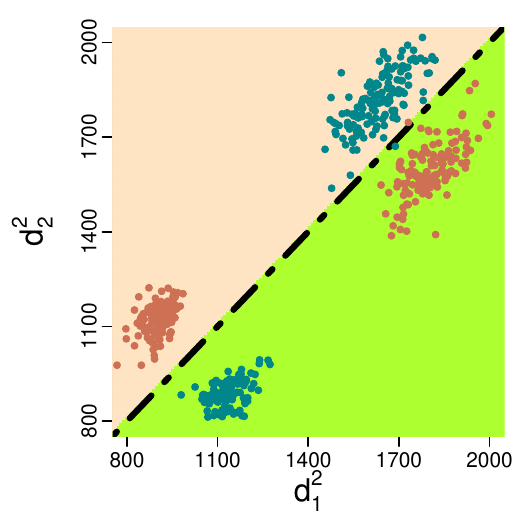} & \includegraphics[width=0.34\linewidth]{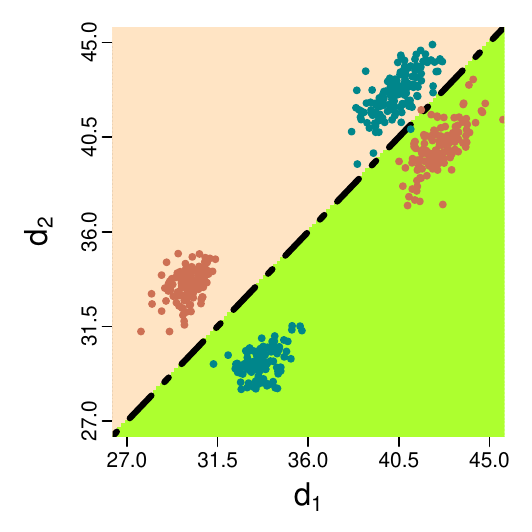} & \includegraphics[width=0.34\linewidth]{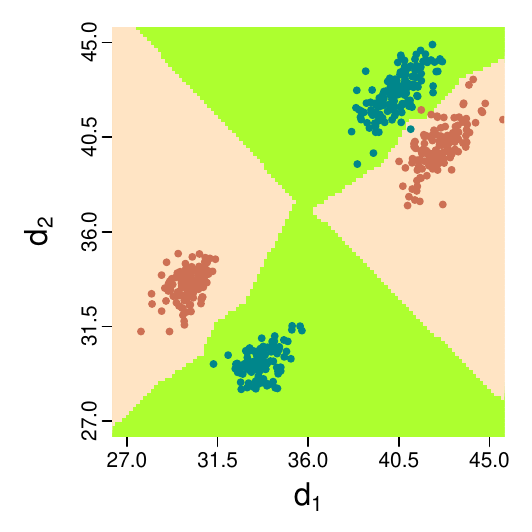} \\ 
\end{tabular}
\caption{Scatter plots of training (top row) and test (bottom row) samples along with the class boundaries estimated by NN, CH, MCH, and MDist classifiers in Example 4.\label{fig:boundary_Ex4}}
\end{figure}

Note that in a binary classification problem with training samples $\{\Xvec_{11},\Xvec_{12},\ldots,\Xvec_{1n_1}\}$ and $\{\Xvec_{21},\Xvec_{22},\ldots,\Xvec_{2n_2}\}$ from the two competing classes, for classification of a test case $\Zvec$, the $1$-NN classifier computes its minimum distances $d_1(\Zvec)=\min_{1 \le i \le n_1} \|\Zvec-\Xvec_{1i}\|$ and $d_2(\Zvec)=\min_{1 \le i \le n_2} \|\Zvec-\Xvec_{2i}\|$ from Class-1 and Class-2, respectively. Then it classifies $\Zvec$ to the first class if $d_2(\Zvec)>d_1(\Zvec)$ or $d_2^2(\Zvec)>d_1^2(\Zvec)$. Like the $1$-NN classifier, the CH classifier also leads to linear classification in the $d_1^2-d_2^2$
space and classifies $\Zvec$ to the first class if $d_2^2(\Zvec)> d_1^2(\Zvec)+C_1$, where $C_1=\frac{1}{2}\big[ \binom{n_2}{2}^{-1} \sum_{s < t}\|\Xvec_{2s}-\Xvec_{2t}\|^2-\binom{n_1}{2}^{-1} \sum_{s < t}\|\Xvec_{1s}-\Xvec_{1t}\|^2$\big].
Similarly, the MCH classifier leads to linear classification in the $d_1-d_2$
space and classifies  $\Zvec$ to the first class if $d_2(\Zvec)> d_1(\Zvec)+C_2$, where $C_2=\frac{1}{2}\big[ \binom{n_2}{2}^{-1} \sum_{s < t}\|\Xvec_{2s}-\Xvec_{2t}\|-\binom{n_1}{2}^{-1} \sum_{s < t}\|\Xvec_{1s}-\Xvec_{1t}\|\big]$. 
The first and the second columns in Figure \ref{fig:boundary_Ex4} show the class boundaries estimated by these classifiers (the back line in the first and the second column shows the class boundary estimated by the $1$-NN classifier) in Example 4 for dimension $500$. They also show the scatter plots of 
$(d_1(\cdot),d_2(\cdot))$ $\big($ or $\big(d_1^2(\cdot),d_2^2(\cdot)\big)\big)$ for all training 
(top row) and test (bottom row) sample observations. For the training data points, the leave-one-out method \citep[see, e.g.,][]{wong2015performance} is used to compute its minimum distances from the two classes. From this figure, it is quite evident that minimum distances (or squared minimum distances) from the two classes contain substantial information about class separability, but the resulting data clouds from the two classes are not linearly separable in that space. As a result, NN, CH, and MCH classifiers, all had poor performance. But we can overcome this problem if we use a suitable nonlinear classifier in that space. For instance one can use the $1$-NN classifier in the $d_1-d_2$
space. This classifier, which is referred to as the MDist classifier, performed well in this example. The last column in Figure \ref{fig:boundary_Ex4} shows the class boundary estimated by the MDist classifier. 
Note that it correctly classified almost all observations.

\begin{figure}[t]
\centering
\setlength{\tabcolsep}{-2pt}
\begin{tabular}{ccc}
(a) NN and CH classifiers & (b) NN and MCH classifiers & (c) MDist classifier \\
\includegraphics[width=0.34\linewidth]{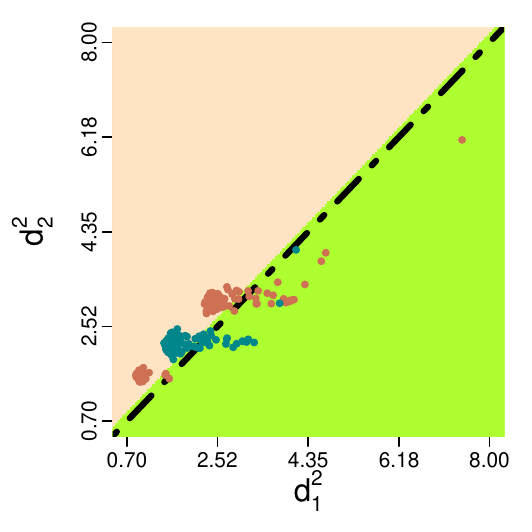} & \includegraphics[width=0.34\linewidth]{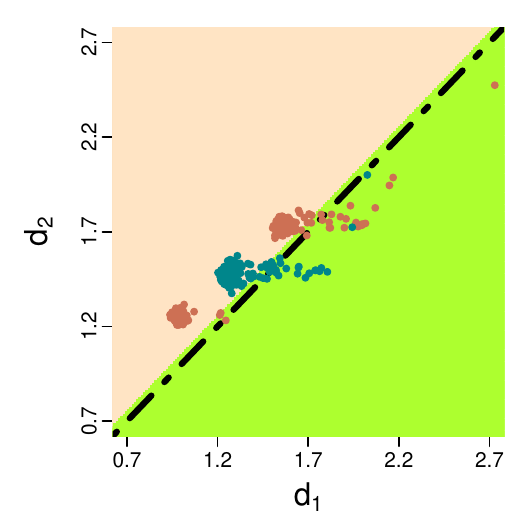} & \includegraphics[width=0.34\linewidth]{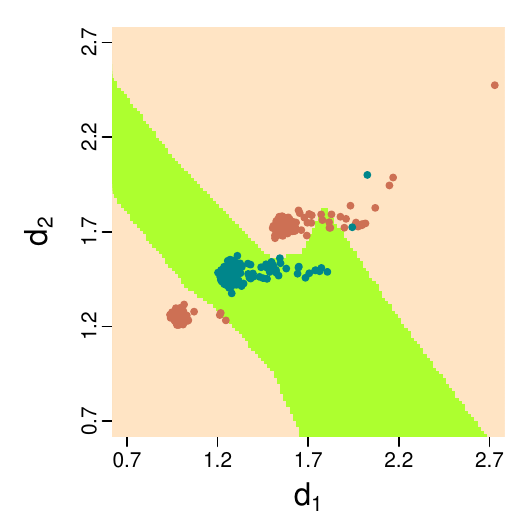} \\   
\end{tabular}
\caption{Scatter plots of the test samples and the class boundaries estimated by NN, CH, MCH and MDist classifiers in Example 5.\label{fig:boundary_Ex5}}
\end{figure}

We observed a similar phenomenon in Example 5 as well (see Figure \ref{fig:boundary_Ex5}).
Like Example 4, here also the observations from different sub-classes form distinct clusters in the $d_1-d_2$
space (or the $d_1^2-d_2^2$ space). So, this feature space contains useful information about class separability, but the feature vectors from the two classes are not linearly separable. Therefore, while 
NN, CH and MCH classifiers had misclassification rates close to 50\%, the MDist classifier had an excellent performance. 

In Example 6, we have the convergence of pairwise distances for observations from the normal distribution, but not for observations from the multivariate $t$ distribution. In this example, NN, CH and MCH classifiers classified almost all observations into a single class (see Figure \ref{fig:boundary_Ex6}), but the MDist classifier
had much superior performance.

\begin{figure}[t]
\centering
\setlength{\tabcolsep}{-2pt}
\begin{tabular}{ccc}
(a) NN and CH classifiers& (b) NN and MCH classifiers & (c) MDist classifier\\
\includegraphics[width=0.34\linewidth]{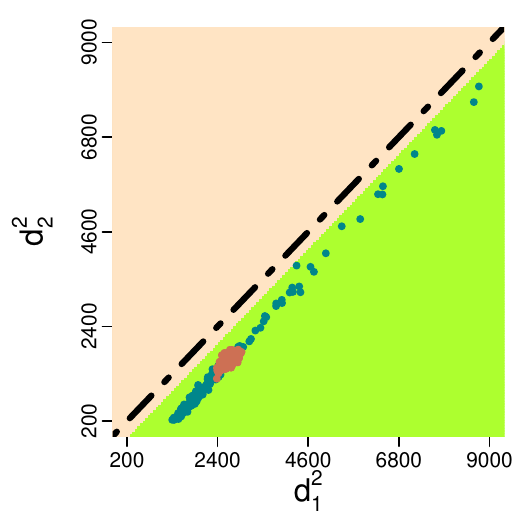} & \includegraphics[width=0.34\linewidth]{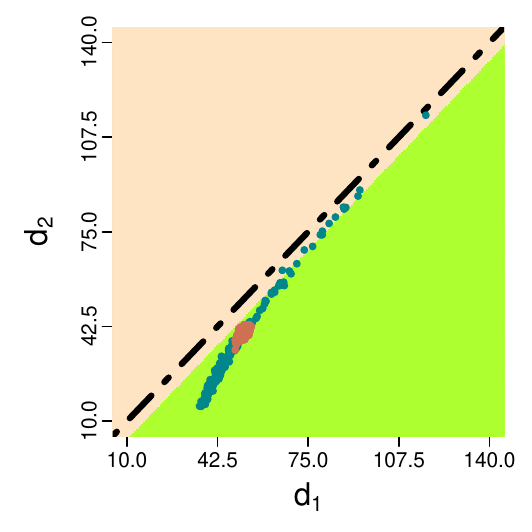} & \includegraphics[width=0.34\linewidth]{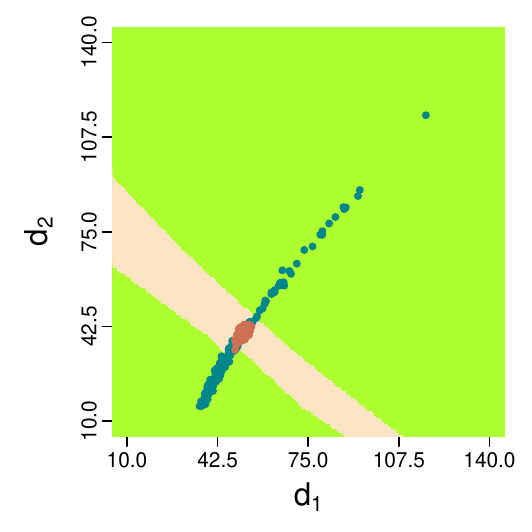} \\   
\end{tabular}
\caption{Scatter plots of the test samples and the class boundaries estimated by NN, CH, MCH and MDist classifiers in Example 6.\label{fig:boundary_Ex6}}
\end{figure}

\begin{figure}[!b]
\centering
\setlength{\tabcolsep}{-2pt}
\begin{tabular}{ccc}
(a) Ex. 4: Two mixture & (b) Ex. 5: Uniform vs. & (c) Ex. 6: Normal$({\bf 0}_d,3{\bf I}_d)$ vs.  \\
 normal distributions & mixture of uniforms & standard $t$ with $3$ d.f.\\
\includegraphics[width=0.34\linewidth]{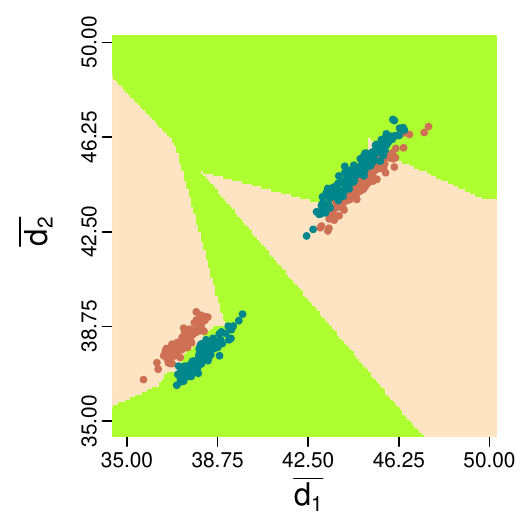} & \includegraphics[width=0.34\linewidth]{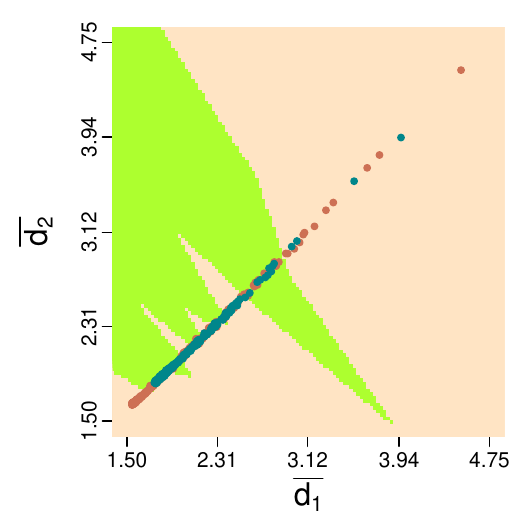} & \includegraphics[width=0.34\linewidth]{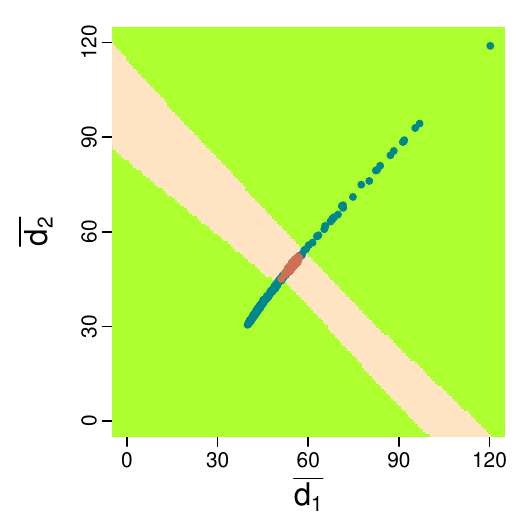} \\   
\end{tabular}
\caption{Scatter plots of the test samples and the class boundaries estimated by the TRAD classifier in Examples 4-6.\label{fig:Boundary_TRAD}}
\end{figure}

\begin{figure}[t]
\centering
\setlength{\tabcolsep}{2pt}
\begin{tabular}{cccc}
(a) Ex. 4: Two mixture & (b) Ex. 5: Uniform vs. & (c) Ex. 6: Normal$({\bf 0}_d,3{\bf I}_d)$ vs.  \\
 normal distributions & mixture of uniforms & standard $t$ with $3$ d.f.\\
\includegraphics[width=0.32\linewidth]{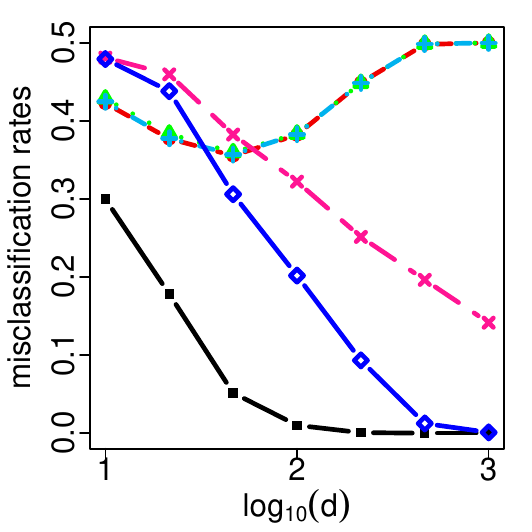} & \includegraphics[width=0.32\linewidth]{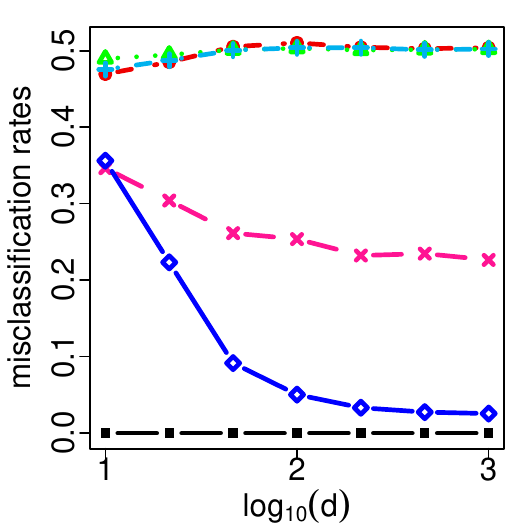} & \includegraphics[width=0.33\linewidth]{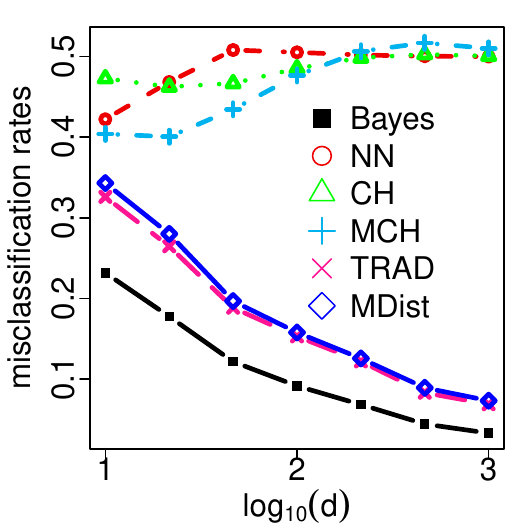} \\ 
\end{tabular}
\caption{Misclassification rates of Bayes, CH, MCH, TRAD and MDist classifiers in Examples 4-6.\label{fig:4-6_2}}
\end{figure}

A similar idea of projecting the observations into a distance-based feature space and using the nearest neighbor classifier on that space was also considered in \cite{dutta2016some}, where the authors suggested using average distances ${\bar d}_1(\Zvec)=\textnormal{avg}_i \|\Zvec-\Xvec_{1i}\|$ and  ${\bar d}_2(\Zvec)=\textnormal{avg}_i \|\Zvec-\Xvec_{2i}\|$ from the competing classes as features. Figure \ref{fig:Boundary_TRAD} shows these features for the test sample observations in Examples 4-6 and also the class boundaries estimated by the resulting classifier, called the TRAD classifier \citep[see][]{dutta2016some}. Here also, for computing the feature vectors for the training sample observations, the leave-one-out method is used. Figure \ref{fig:Boundary_TRAD}(a) shows that in Example 4, we have reasonable separability in the feature space, but the four distinct clusters are not as prominent as they were in Figure \ref{fig:boundary_Ex4}. Here we have some overlaps between the clusters corresponding to two competing classes. As a result, TRAD performed better than NN, CH and MCH classifiers, but not as good as the MDist classifier. This is also evident from 
Figure \ref{fig:4-6_2}(a), which shows the average (over 100 replications) test set misclassification rates of these classifiers for various choices of $d$. In Example 5, the features based on average distances
do not provide much separation between the two classes (see Figure \ref{fig:Boundary_TRAD}(b)). So,
as expected, TRAD had much higher misclassification rates (see \ref{fig:4-6_2}(b). However, in 
Example 6, the ${\bar d}_1-{\bar d}_2$ space shows almost the same degree of separation
as in the $d_1-d_2$ space (see Figure \ref{fig:Boundary_TRAD}).
So, the class boundaries estimated by TRAD and MDist classifiers were almost similar, and they had almost similar misclassification rates. (see Figures \ref{fig:4-6_2}(c)).

The success of the MDist classifier motivates us to carry out some theoretical analysis to understand its high-dimensional behavior. For this investigation, we again consider
assumptions (A1)-(A3), and prove the perfect classification property of the MDist classifier in high dimensions.

\begin{theorem}
Suppose that $J$ competing classes satisfy assumptions (A1)-(A3), and from each of them, there are at least two observations (i.e, $n_j\ge 2$ for all $j=1,2,\ldots, J$). If 
for all $j \neq i$, $\nu^2_{ji}>0$ or $\sigma_j^2 \neq \sigma_i^2$, the misclassification rate of the MDist classifier converges to $0$ as $d$ grows to infinity.
\end{theorem}

However, as we have discussed before, if the competing classes are mixtures of several sub-classes, the assumptions (A1)-(A3) may hold for each of the sub-classes but none of the competing classes (as in Examples 4). We have seen that in such situations, CH and MCH classifiers often have poor performance in high dimensions. However, from the proof of Theorem 2 (see Appendix), it is clear that in such cases, for each of the sub-classes, the feature vectors
of minimum distances converge to a point as the dimension increases. If these points are distinct for each sub-class,
we get some distinct clusters in the feature space, and the MDist classifier leads to perfect classification. We have already seen that in Example 4. A theorem similar to Theorem 2 can be stated for these mixture distributions as well, but the conditions for perfect classification by the MDist classifier (i.e., the 
conditions needed to ensure that for any two sub-classes from two competing classes, the feature vectors converge to two distinct points, one for each sub-class) becomes mathematically complicated to interpret. That is why we choose not to state that theorem here.

Now we consider two interesting examples (Example 7 and 8) involving binary classification, where each of the two competing classes satisfies assumptions (A1)-(A3), but we have $\nu_{12}^2=0$ and $\sigma_1^2=\sigma_2^2$. So, in this case, the feature vectors
$(d_1(\cdot),d_2(\cdot))$ corresponding to two competing classes 
converge to the same limiting value as $d$ increases. One may be curious to know how the MDist classifier performs in such situations, and we investigate it here. Here also, we consider different values of $d$ ranging between $10$ and $1000$, form the training and the test sets of size 50 and 500 by taking an equal number of observations from the two classes and repeat the experiment $100$ times to compute the average test set misclassification rates of different classifiers.


\begin{figure}[!b]
\centering
\setlength{\tabcolsep}{-2pt}
\begin{tabular}{cccc}
(a) MDist classifier (Ex. 7)& (b) TRAD classifier (Ex. 7) & (c) MDist1 classifier (Ex. 7)\\
\includegraphics[width=0.33\linewidth]{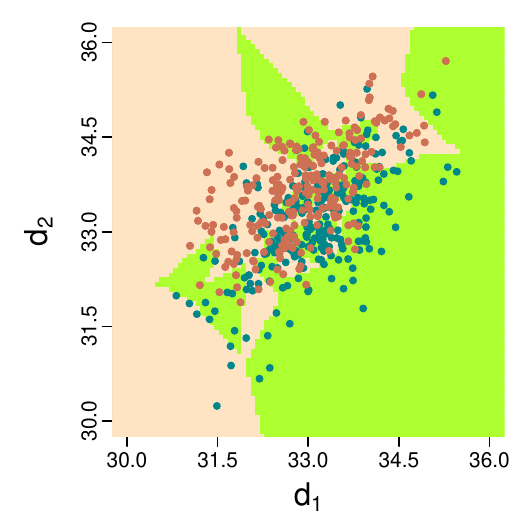} & \includegraphics[width=0.33\linewidth]{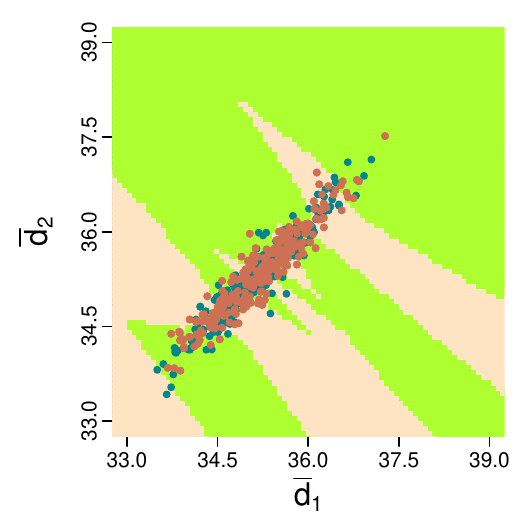} & \includegraphics[width=0.33\linewidth]{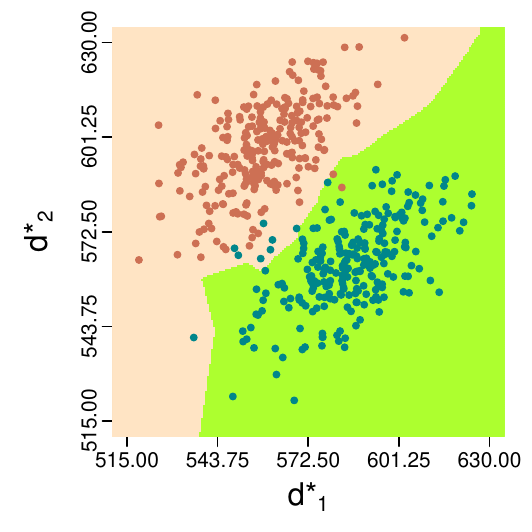} &  \\ 
\vspace{-0.05in}
(d) MDist classifier (Ex. 8)& (e) TRAD classifier (Ex. 8) & (f) MDist1 classifier (Ex. 8)\\
\includegraphics[width=0.33\linewidth]{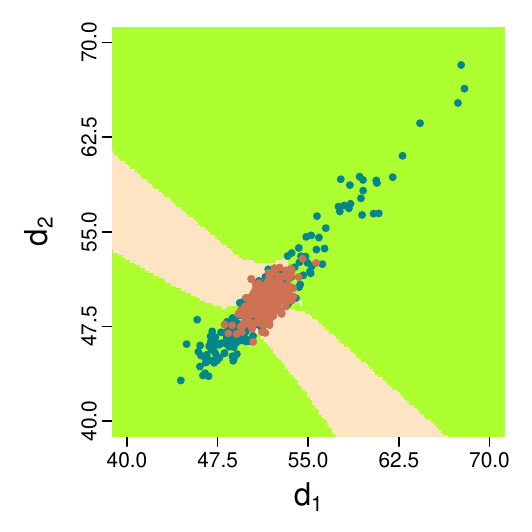} & \includegraphics[width=0.33\linewidth]{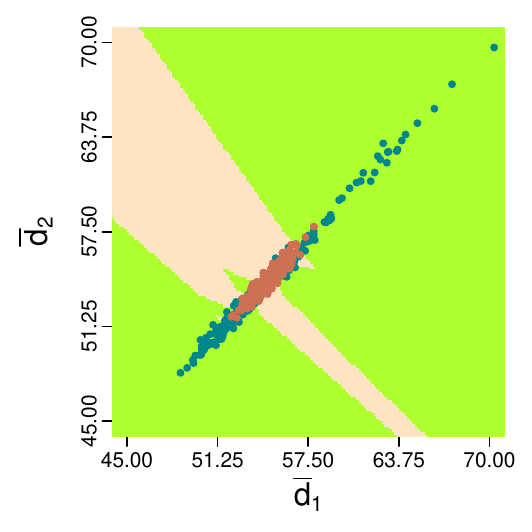} & \includegraphics[width=0.33\linewidth]{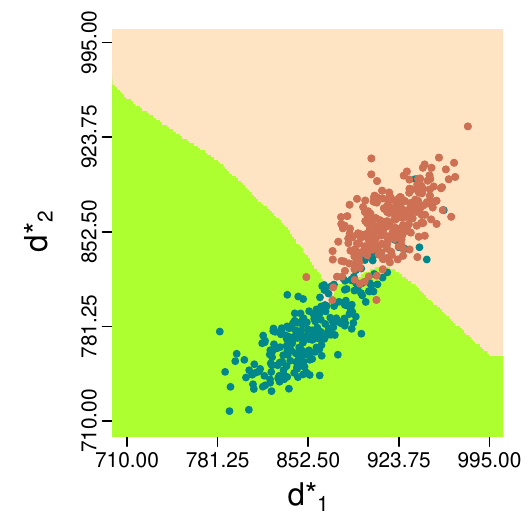} &  \\  
\end{tabular}
\caption{Scatter plots of the test samples and the class boundaries estimated by the MDist, TRAD and MDist1 classifiers in Example 7 (top row) and Example 8 (bottom row).\label{fig:Boundary_7-8}}
\end{figure}

{\textbf{Example 7:} {\it We consider two normal distributions having the same mean vector ${\bf 0}_d$ but different dispersion matrices $\Lambda_1$=diag$(\lambda_{11},\ldots,\lambda_{1d})$ and $\Lambda_2$=diag$(\lambda_{21},\ldots,\lambda_{2d})$. 
Here $\lambda_{1i}=1/2$ and $\lambda_{2i}=2$ for $i\le d/2$, whereas  $\lambda_{1i}=2$ and $\lambda_{2i}=1/2$ for $i > d/2$.}

Figure \ref{fig:Boundary_7-8}(a) show the scatter plots of the test set observations in the $d_1-d_2$ space and the class boundary estimated by the MDist classifier for $d=500$. It is clear that unlike previous examples, here the features based on minimum distances fail to discriminate between the two classes. As a result, the MDist classifier had much higher misclassification rates (see Figure \ref{fig:7-8}(a)).
The features based on average $\ell_2$ distances also fail to provide any discriminatory information (see Figure \ref{fig:Boundary_7-8}(b)) in this case. The performance of the TRAD classifier was even worse. It misclassified almost half of the test set observations (see Figure \ref{fig:7-8}(a)). Surprisingly, in this example, we
get a good result if instead of $\ell_2$ distance (Euclidean distance), we
use $\ell_1$ distance (Manhattan distance) for finding the neighbors. If $\{\Xvec_{11},\Xvec_{12},\ldots,\Xvec_{1n_1}\}$ and $\{\Xvec_{21},\Xvec_{22},\ldots,\Xvec_{2n_2}\}$ are training sample observations from two competing classes (here we have $n_1=n_2=25$), for
any $\Zvec$, we can use $d_1^{\ast}(\Zvec)=\min_{1 \le i \le n_1} \|\Zvec-\Xvec_{1i}\|_1$ and
$d_2^{\ast}(\Zvec)=\min_{1 \le i \le n_2} \|\Zvec-\Xvec_{2i}\|_1$ as features and perform usual nearest neighbor classification on that feature space. Here also for computing $d_1^{\ast}$ and $d_2^{\ast}$ at the training data points, we use the leave-one-out method. Figure \ref{fig:Boundary_7-8}(c) shows the scatter plot of these features for all test set observations and the class boundary estimated by the $1$-NN classifier on this feature space 
(we call it the MDist1 classifier). Here we have two distinct clusters in the feature space, one for each class. As a result, the MDist1 classifier
had an excellent performance and correctly classified almost all observations. The average test set misclassification rate of this classifier (reported in  Figure \ref{fig:7-8}(a)) also tells us the same story. {Now, let us consider the following example. }

{\textbf{Example 8:} {\it Here each of the two classes has i.i.d. measurement variables. In Class-1, they follow the $N(0,3)$ distribution, while in Class-2, they follow the standard $t$ distribution with $3$ degrees of freedom.}

Note that this is different from the multivariate $t$ distribution considered in Example 6. In this example also, the features based
on minimum $\ell_2$ distances and those based on average $\ell_2$ distances
do not provide much separability
between the two classes (see Figure \ref{fig:Boundary_7-8}(d) and (e)), but the features based on minimum $\ell_1$ distances make the data clouds better separated (see Figure \ref{fig:Boundary_7-8}(f)). As a result, the MDist1 classifier outperformed TRAD and MDist classifiers (see Figure \ref{fig:7-8}(b)).

\begin{figure}[t]
\centering
\begin{tabular}{c@{\hskip -20pt}c@{\hskip 2pt}c}
(a) Ex. 7: Two normals with different & & (b) Ex. 8: Product of $N(0,3)$ vs \\
dispersion matrices having same trace & & product of univariate $t_3$\\
\includegraphics[height=5.50cm, width=0.38\linewidth]{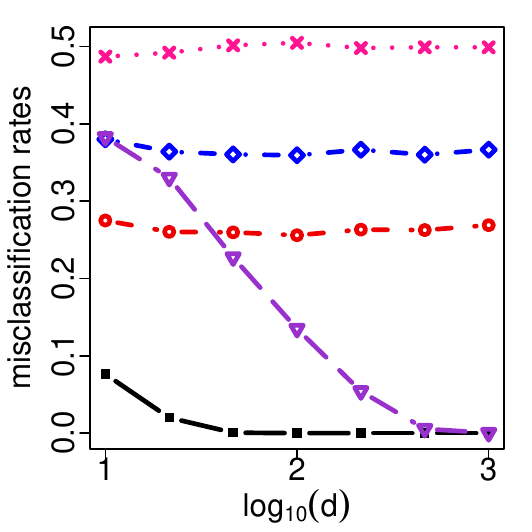} & \includegraphics[height=5.50cm, width=0.22\linewidth]{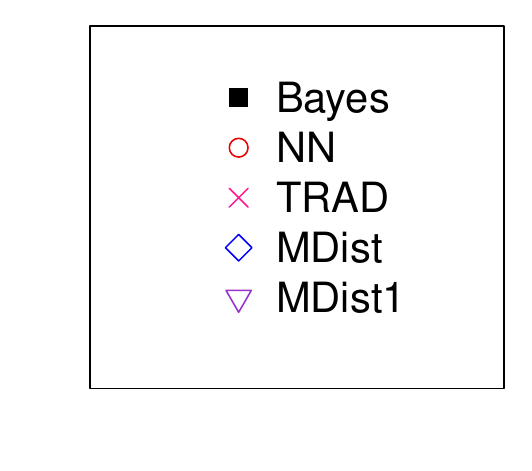} &
\includegraphics[height=5.50cm, width=0.38\linewidth]{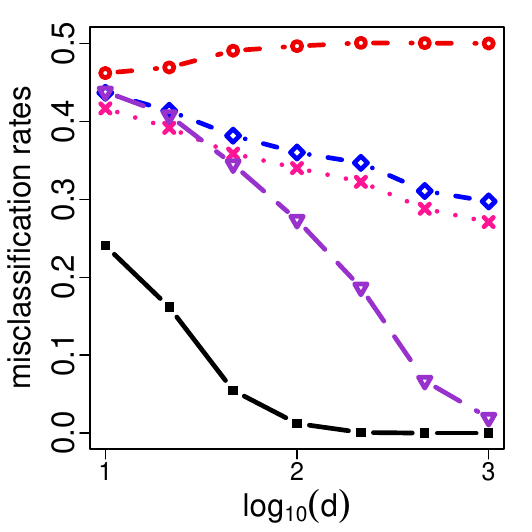} \\ 
\end{tabular}
\caption{Misclassification rates of Bayes, NN, TRAD, MDist and MDist1 classifiers in Examples 7 and 8.\label{fig:7-8}}
\end{figure}


To understand the high-dimensional behavior of the MDist1 classifier, we carry out some theoretical investigations under the following assumptions, which are similar to (A1)-(A3) stated before. 

\begin{enumerate}
    \item[(A1$^{\circ}$)] In each of the $J$ competing classes, the measurement variables have uniformly bounded second moments.
    \item[(A2$^{\circ}$)] If $\Xvec = (X_1,\ldots,X_d)^{\top}\sim F_j$ and $\Yvec = (Y_1,\ldots,Y_d)^{\top}\sim F_i$ ($1\le j,i\le J$) are independent, for ${\bf U}=\Xvec-\Yvec$,  $\sum_{r \neq s}|Corr(|U_r|,|U_s|)|$
    is of the order $o(d^2)$.
\item[(A3$^{\circ}$)] For independent random vectors $\Xvec \sim F_j$ and $\Yvec \sim F_i$ ($1\le j,i\le J)$, $E\big(\frac{1}{d}\|\Xvec-\Yvec\|_1\big) = \frac{1}{d}\sum_{q=1}^{d}E|X_q-Y_q|$ converges to a constant $\tau_{ji}$ as $d \rightarrow \infty$.
\end{enumerate} 

Under (A1$^{\circ}$) and (A2$^{\circ}$), we have the convergence of pairwise $\ell_1$ distances. Following similar steps as used in the proof of Lemma 1, one can show that for $\Xvec\sim F_j$ and $\Yvec \sim F_i$ ($1 \le j,i\le J$), $\big|\frac{1}{d}\|\Xvec-\Yvec\|_1 -E\big(\frac{1}{d}\|\Xvec-\Yvec\|_1\big)\big| \stackrel{P}{\rightarrow}0$ as $d \rightarrow \infty$. 

For any $q=1,2,\ldots d$, define 
$e_{ji}^{(q)}=2E|X_q-Y_q|-E|X_q-X_q^{\prime}|-E|Y_q-Y_q^{\prime}|$, where $\Xvec, \Xvec^{\prime}\sim F_j$ and 
$\Yvec, \Yvec^{\prime}\sim F_i$
($j \neq i$) are independent random vectors. This quantity $e_{ji}^{(q)}$ can be viewed as the energy distance \citep[see, e.g.,][]{szekely2023energy} between the $q$-th marginals of $F_j$ and $F_i$ ($F_j^{(q)}$ and $F_i^{(q)}$, say). Following \cite{baringhaus2004new}, one can show that $e_{ji}^{(q)}$ is non-negative, and it takes the value $0$ if and only if $F_j^{(q)}=F_i^{(q)}$. So, for any fixed dimension $d$, we have 
$\frac{1}{d}\big[2E\|\Xvec-\Yvec\|_1 - E\|\Xvec-\Xvec^{\prime}\|_1 - E\|\Yvec-\Yvec^{\prime}\|_1\big]
 = \frac{1}{d} \sum_{q=1}^{d} e_{ji}^{(q)}
= {\bar e}_{ji}(d) \ge 0$, where the equality holds if and only if
$F_j^{(q)}=F_i^{(q)}$ for $q=1,2,\ldots,d$. Therefore, it is somewhat reasonable to assume
that ${\cal E}_{ji}=\lim_{d \rightarrow \infty} 
{\bar e}_{ji}(d)>0$, which essentially says that the average coordinate-wise energy distance is asymptotically non-negligible. Under this assumption, we have the perfect separation property of the MDist1 classifier, which is asserted by the following theorem.

\begin{theorem}
Suppose that $J$ competing classes satisfy assumptions (A1$^{\circ}$)-(A3$^{\circ}$), and from each of them, there are at least two observations (i.e, $n_j\ge 2$ for all $j=1,2,\ldots, J$). If the limiting value of the average coordinate-wise energy distance ${\cal E}_{ji}>0$ 
for all $j \neq i$, the misclassification rate of the MDist1 classifier converges to $0$ as $d$ grows to infinity. 
\end{theorem}

Note that while TRAD and MDist classifiers fail to discriminate between two distributions differing outside the first two moments, the MDist1 classifier
can discriminate between them as long as they differ in their one-dimensional marginals. That is why it outperformed TRAD and MDist classifiers in Examples 7 and 8. 

This classifier enjoys the perfect separation property in high dimensions even when the competing classes are mixtures of several sub-classes, and these sub-class distributions satisfy assumptions (A1$^{\circ}$)-(A3$^{\circ}$). It becomes clear from the proof of Theorem 3 (see Appendix) that in such cases, for each of the sub-classes, the feature vectors
of minimum $\ell_1$ distances  (after appropriate scaling) converge to a point as the dimension increases. If these points are distinct for each sub-class, the MDist1 classifier leads to perfect classification. But here also writing the conditions for perfect classification becomes mathematically complicated to interpret. So, we decide not to state another theorem in this regard. However, our analysis of simulated data sets clearly demonstrates this. Figure \ref{fig:1-6} shows the misclassification rates of TRAD, MDist and MDist1  classifiers in Examples 1-6 along with those of $1$-NN and Bayes classifiers. In all these examples including Example 4 and 5, where we deal with mixture distributions, MDist and MDist1 classifiers had similar performance. 

This figure shows another interesting phenomenon.
In Examples 1-3, when the underlying distributions are unimodal, the TRAD classifier, which considers the average of distances from all observations, performed better than MDist and MDist1 classifiers, which consider the distance of one nearest neighbor only. In Example 6 also, TRAD had an edge over the other two classifiers. But, in the case of mixture distributions (see Examples 4 and 5), taking the average over all observations coming from different sub-classes does not seem to be a meaningful option. In those cases, MDist and MDist1 classifiers outperformed the TRAD classifier. These result shows that instead of always going for features based on a single nearest neighbor from each class, sometimes it is more meaningful to consider distances from multiple nearest neighbors. We can include these distances in the set of features and go for classification in the extended feature space. We consider such methods in the following subsection.

\begin{figure}[h]
\centering
\setlength{\tabcolsep}{-2pt}
\begin{tabular}{cccc}
(a) Example 1 & (b) Example 2 & (c) Example 3 \\
\includegraphics[width=0.32\linewidth]{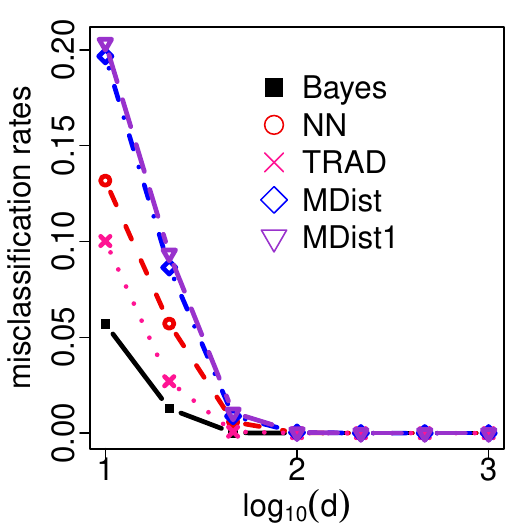} & \includegraphics[width=0.32\linewidth]{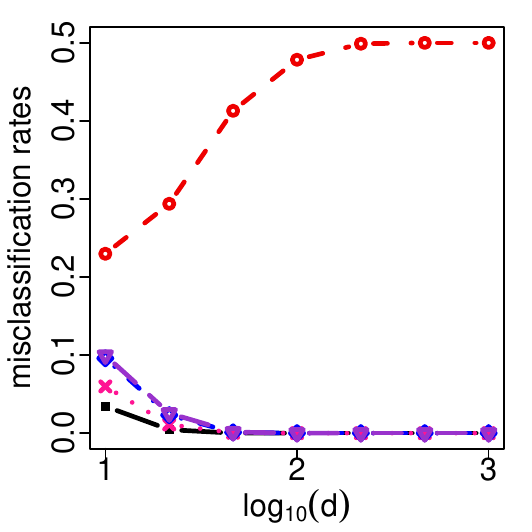} & \includegraphics[width=0.32\linewidth]{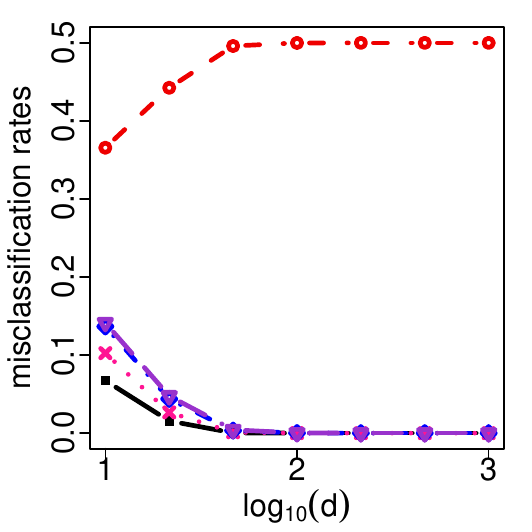} \\  
(d) Example 4 & (e) Example 5 & (f) Example 6 \\
\includegraphics[width=0.32\linewidth]{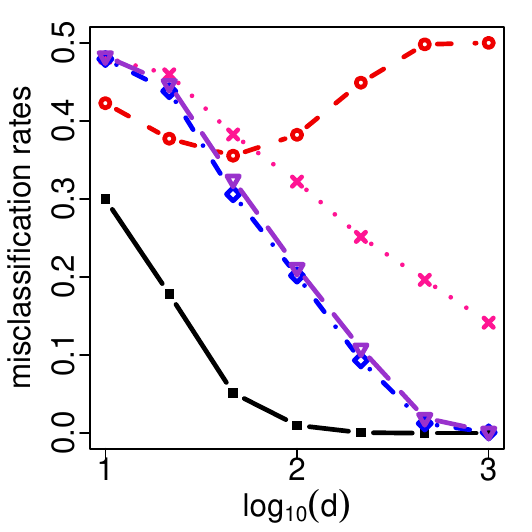} & \includegraphics[width=0.32\linewidth]{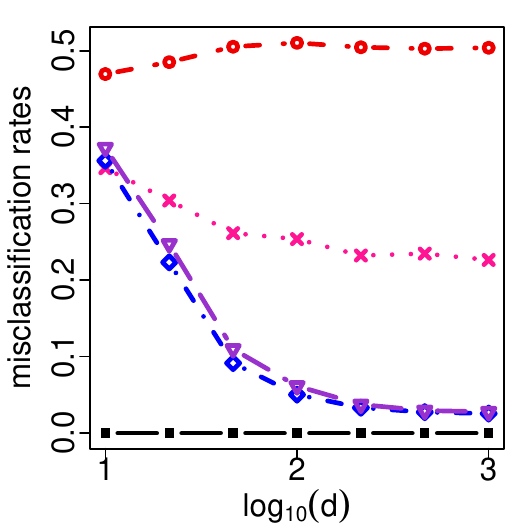} & \includegraphics[width=0.32\linewidth]{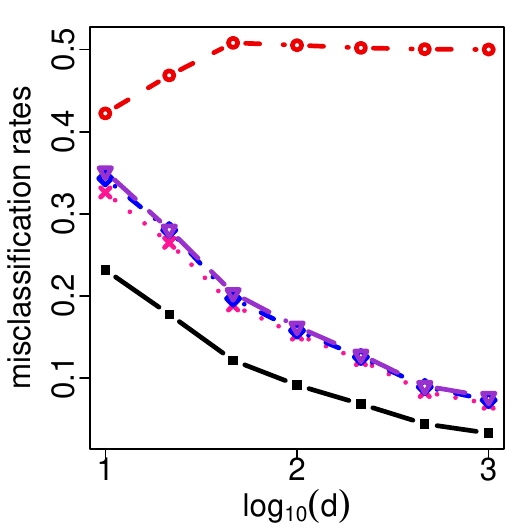} \\  
\end{tabular}
\caption{Misclassification rates of Bayes, NN, TRAD, MDist and MDist1 classifiers in Examples 1-6.\label{fig:1-6}}
\vspace{0.1in}
\end{figure}

\subsection{Clssification based on multiple neighbors}

\begin{figure}[!b]
\centering
\setlength{\tabcolsep}{-2pt}
\begin{tabular}{cccc}
(a) Example 1 &
(b) Example 2 &
(c) Example 3 \\
\includegraphics[width=0.32\linewidth]{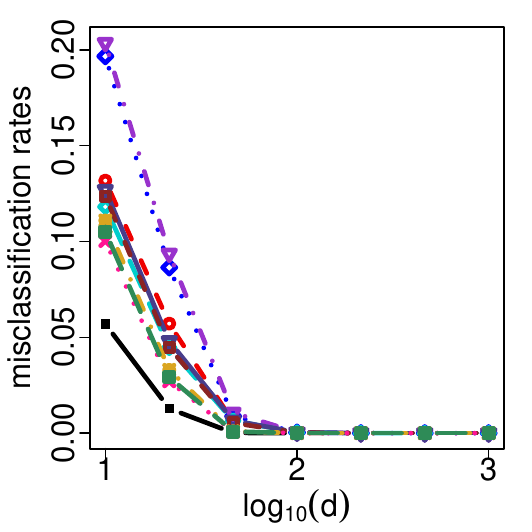} & \includegraphics[width=0.32\linewidth]{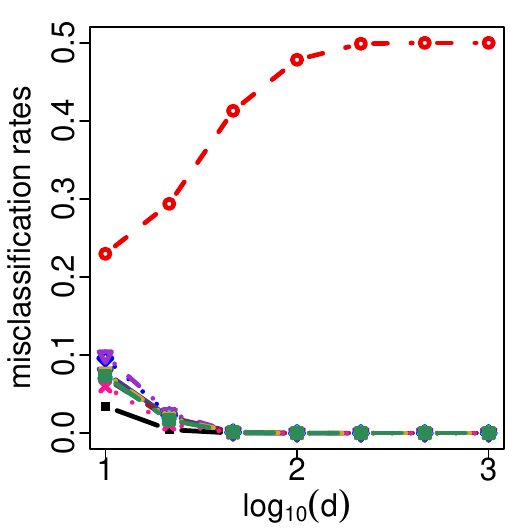} & \includegraphics[width=0.32\linewidth]{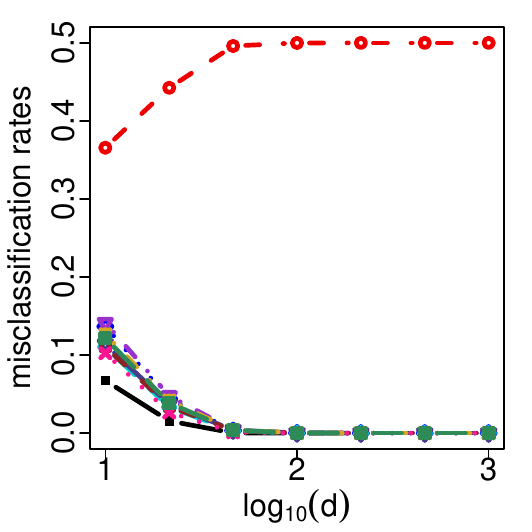} \\ 
(d) Example 4 &
&
(e) Example 5 \\
\includegraphics[width=0.32\linewidth]{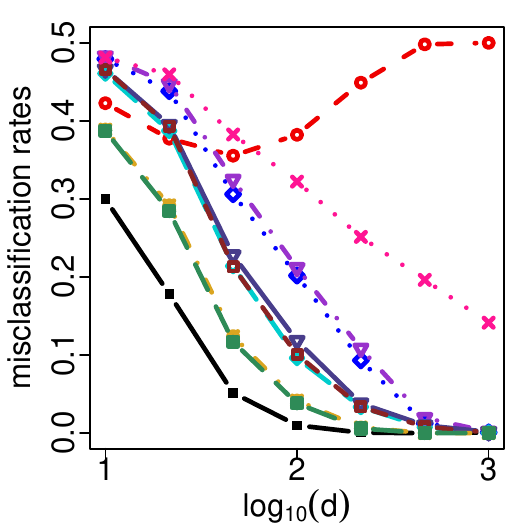} & 
\includegraphics[width=0.28\linewidth]{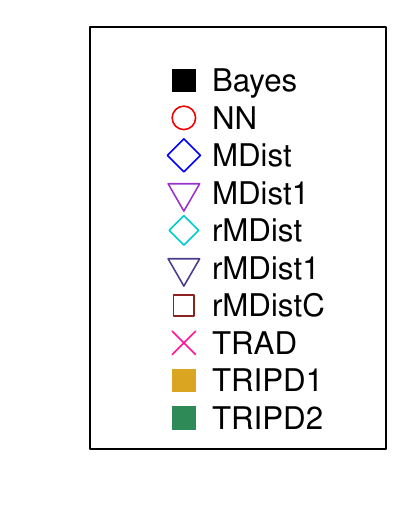} & \includegraphics[width=0.32\linewidth]{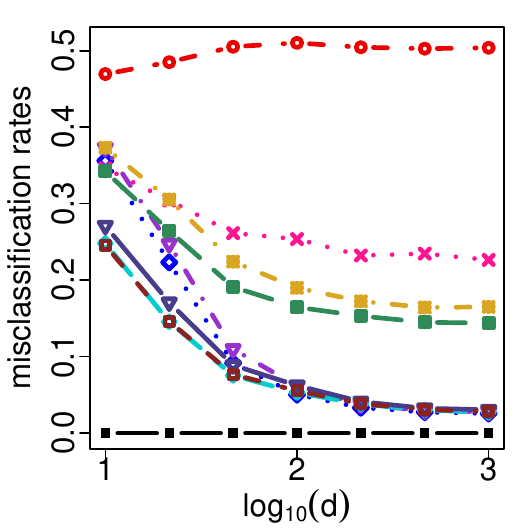} \\
(f) Example 6 &
(g) Example 7 &
(h) Example 8 \\
\includegraphics[width=0.32\linewidth]{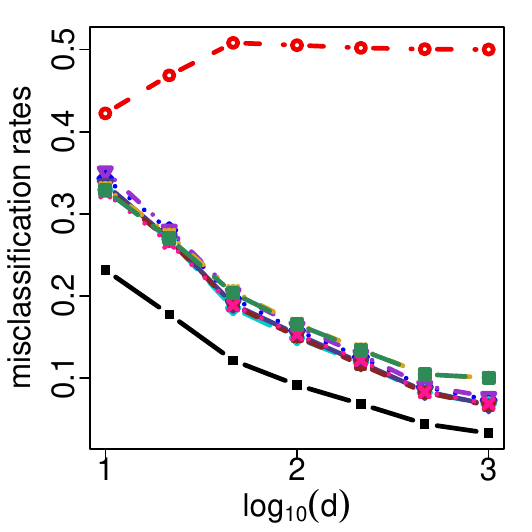} & 
\includegraphics[width=0.32\linewidth]{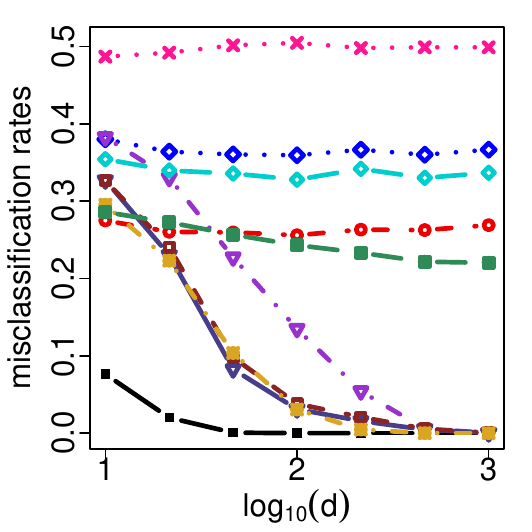} & 
\includegraphics[width=0.32\linewidth]{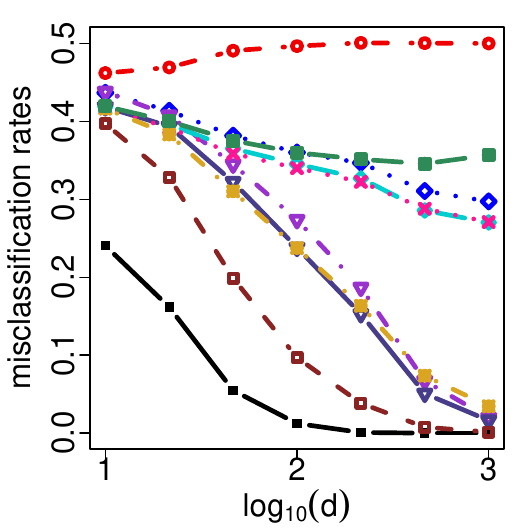} \\  
\end{tabular}
\caption{Misclassification rates of different 
classifiers in Examples 1-8.\label{fig:1-9}}
\vspace{0.1in}
\end{figure}

Instead of considering only the distance of the first neighbor from each class, here we consider the distances of the first $r$ ($r\ge 1$) neighbors from each class and use them as features. So, if there are $J$ competing classes, we consider a total of $Jr$ many features and use the $1$-NN classifier on that feature space. Here also, we can use $\ell_2$ distances or $\ell_1$ distances as features, and the resulting classifier is referred to as rMDist and rMDist1 classifiers, respectively. One may also consider both $\ell_1$ and $\ell_2$ distances and deal with $2Jr$ many features simultaneously. We refer to the resulting classifier as the rMDistC classifier. In all these cases the value of $r$ is chosen by minimizing the leave-one-out cross-validation estimate \citep[see, e.g.][]{wong2015performance} of the misclassification rate. Figure \ref{fig:1-9} shows the average test set misclassification rates of these classifiers in Examples 1-8. One can see that in most of the examples, rMDist and rMDist1 classifiers performed better than MDist and MDist1 classifiers, respectively. The rMDistC classifier also performed well in almost all examples. In Examples 7 and 8, it outperformed rMDist and rMDist1 classifiers.

 A somewhat similar classification method was considered in \cite{dutta2016some}, where the authors transformed each observation
into an $n$-dimensional vector containing its distances from all training sample observations. They also considered $\ell_1$ and $\ell_2$
distances for transformation and called the resulting classifiers as
TRIPD1 and TRIPD2. respectively. Misclassification rates of those two methods are also reported in Figure \ref{fig:1-9}. In Examples 1-3 and 6, their misclassification rates were similar to our proposed methods. In Example 4, they had the lowest misclassification rates, but in Example 5, they were outperformed by our proposed classifiers. In Examples 7 and 8, the TRIPD2 classifier, which is based on $\ell_2$ distances, had higher misclassification rates. In Example 7, the performance of the TRIPD1 classifier was comparable to rMDist1 and rMDistC classifiers, but in Example 8, the rMdistC classifier had a clear edge.  

\section{Results from the analysis of  benchmark datasets}

 We analyze $10$ benchmark datasets for further evaluation of the performance of the proposed and existing methods discussed in the previous two sections. For these benchmark datasets, since the true class distributions are not known, it is not possible to compute the Bayes risks. Therefore, to facilitate comparison, we report the misclassification rates of two popular classifiers, support vector machines (SVM) \citep[see, e.g.][]{christianini2003support, scholkopf2018learning} and random forest (RF) \citep[see, e.g.][]{breiman2001random,genuer2020random}, which are known to perform well for high dimensional data.
 Since the nearest neighbor classifiers are nonlinear, to make it fair, here we use the nonlinear SVM for comparison. For our numerical study, we use the radial basis function kernel, where all tuning parameters are chosen using the 5-fold cross-validation method \citep[see, e.g.,][]{wong2015performance}. 
We use the R package ${\tt caret}$ for this purpose. The same package is used for the random forest classifier as well, where we use default tuning parameters.

\begin{table}[!t]
    \centering
    \small
    \begin{tabular}{|ccccc |c| ccccc|} \cline{1-5} \cline{7-11}
     dataset & $d$ & $J$& \multicolumn{2}{c|}{Sample size} &  & dataset & $d$ & $J$& \multicolumn{2}{c|}{Sample size} \\
     &  & & Train & Test & & &  & & Train & Test\\ \cline{1-5} \cline{7-11}

      Synthetic Control & 60 & 6 & 60 & 540 & &
          Lightning7 & 319 & 7 & 70 & 73\\
      \cline{1-5} \cline{7-11}

       Chowdary & 182 & 2 & 52 & 52 &  &
 Herring & 512 & 2 & 64 & 64\\
       \cline{1-5} \cline{7-11}

        
      Trace & 275 & 4 & 100 & 100 & &
             Nutt & 1070 & 2 & 14 & 14\\
        \cline{1-5} \cline{7-11}

        Toe Segmentation1 & 277 & 2 & 40 & 228  & & 
         Gordon & 1628 & 2 & 90 & 91 \\
        \cline{1-5} \cline{7-11}

        Coffee & 286 & 2 & 28 & 28 & & 
       Colon Cancer & 2000 & 2 & 31 & 31  \\ 
          \cline{1-5} \cline{7-11}
    
    \end{tabular}
    
    \caption{Brief descriptions of the benchmark datasets.\label{tab:datasets}}
    \label{tab:1}

\vspace{0.5in}

    \setlength{\tabcolsep}{2pt}
    \small
    \begin{tabular}{|c|cccccccccc|} \hline
Dataset & Synth. & Chowdary & Trace & Toe Seg- & Coffee & Lightning7 & Herring & Nutt & Gordon & Colon \\
 & Control & &&  ment.1 &  & & & & & Cancer \\ \hline 
NN & 18.78 & 4.83 & 20.33 & 38.62 & 2.00 & 37.97 & 51.33 & 34.00 & 2.96 & 26.10\\
& (0.28) & (0.29) & (0.37) & (0.36) & (0.31) & (0.43) & (0.59) & (0.92) & (0.16) & (0.65)\\ \hline
MDist & 10.02 & 7.98 & 13.51 & 42.80 & 2.61 & 39.29 & 45.53 & 14.14 & 1.92 & 32.42 \\
& (0.26) & (0.45) & (0.46) & (0.45) & (0.34) & (0.47) & (0.52) & (0.74) & (0.17) & (0.95)\\
MDist1 & 12.29 & 7.04 & 18.88 & 37.30 & 4.43 & 35.72 & 47.20 & 15.71 & 1.19 & 34.74\\
& (0.30) & (0.46) & (0.46) & (0.45) & (0.39) & (0.52) & (0.57) & (0.70) & (0.10) & (0.94)\\
rMDist & 10.06 & 6.56 & 14.90 & 40.92 & 2.93 & 38.16 & 46.86 & 15.14 & 1.73 & 22.06\\
&(0.28) & (0.37) & (0.48) & (0.38) & (0.32) & (0.42) & (0.54) & (0.86) & (0.14) & (0.92) \\
rMDist1 & 10.12 & 5.35 & 19.61 & 35.64 & 4.50 & 36.00 & 46.06 & 15.93 & 1.27 & 27.03\\
& (0.27) & (0.35) & (0.44) & (0.47) & (0.39) & (0.44) & (0.63) & (0.68) & (0.10) & (1.03)\\
rMDistC & 9.25 & 5.90 & 15.01 & 35.61 & 3.07 & 34.67 & 46.88 & 14.21 & 1.64 & 24.65\\
& (0.27) & (0.32) & (0.49) & (0.47) & (0.34) & (0.47) & (0.54) & (0.80) & (0.15) & (0.96)\\ \hline
TRAD & 14.78 & 7.31 & 24.48 & 49.93 & 4.11 & 37.28 & 48.08 & 12.07 & 6.52 & 18.06\\
& (0.30) & (0.34) & (0.37) & (0.41) & (0.43) & (0.39) & (0.56) & (0.72) & (0.19) & (0.72)\\
TRIPD1 & 7.67 & 4.42 & 23.25 & 33.92 & 6.14 & 31.47 & 47.36 & 17.57 & 1.21 & 25.77\\
& (0.19) & (0.25) & (0.43) & (0.32) & (0.40) & (0.39) & (0.53) & (0.95) & (0.10) & (0.66)\\ 
TRIPD2 & 6.42 & 5.38 & 21.08 & 38.15 & 3.79 & 32.27 & 50.50 & 8.93 & 3.38 & 21.58\\
& (0.19) & (0.30) & (0.39) & (0.34) & (0.39) & (0.40) & (0.56) & (0.86) & (0.20) & (0.62)\\ \hline
Random & 13.07 & 5.00 & 14.23 & 38.21 & 3.21 & 28.29 & 40.30 & 21.07 & 0.93 & 29.55\\
Forest & (0.24) & (0.26) & (0.48) & (0.41) & (0.45) & (0.47) & (0.47) & (1.09) & (0.11) & (0.64) \\ \hline
Nonlin.&9.50 & 7.63 & 10.48 & 45.02 & 4.25 & 35.77 & 39.11 & 11.21 & 2.53 & 20.90 \\ 
SVM & (0.32) & (0.50) & (0.36) & (0.37) & (0.46) & (0.47) & (0.41) & (0.75) & (0.22) & (0.75)\\ \hline
 
    \end{tabular}
    
    \caption{Average misclassification rates (in \%) of {{different classifiers and their standard errors (reported inside the bracket)}} in benchmark datasets.\label{tab:datasets}}
\end{table}

Out of these 10 datasets, Chowdary and Nutt datasets are taken from \href{https://schlieplab.org/Static/Supplements/CompCancer/datasets.htm}{CompCancer dataset-Schliep lab}. The rest of the datasets are taken from the 
\href{http://www.cs.ucr.edu/~eamonn/time_series_data/}{UCR Time Series Classification Archive}. Detailed descriptions of these datasets are available at these respective sources. The datasets taken from the UCR archive have specific training and test sets. We merge these two sets and divide the pooled dataset randomly into two parts to form the training and the test samples. Except for the Synthetic Control Chart data,
in all other cases, the sizes of training and test samples are taken to be the same as they are in the data archive. 
Note that in all these cases, the size of the training sample is smaller than the dimension. In the case of Synthetic Control Chart data, instead of an equal partition (as in UCR achieve), we use training and test samples of size 60 and 540, respectively, so that the training sample size does not 
{become} larger than the dimension.
The datasets from the CompCancer database do not have specific training and test samples. In these cases, we divide the data sets into equal halves to form the training and the test samples. Brief descriptions of these datasets are given in Table \ref{tab:1}. In all these cases, we form the training and the test samples in such a way that the proportions of different classes in the two samples are as close as possible. In each case, this partitioning is carried out $100$ times, and the average test set misclassification rates of different classifiers are reported in Table \ref{tab:datasets}
{{along with their corresponding standard errors}}.
Overall performances of CH and MCH
classifiers (especially, that of the former one)  were much inferior compared to all other classifiers considered here. Therefore, we do not report them in this section.

Though the $1$-NN classifier had the lowest misclassification rate in the Coffee dataset and the second lowest misclassification rate in the Chowdary dataset, in many cases, its performance was far from the best one (see Table \ref{tab:datasets}). For instance, in Nutt and Synthetic Control Chart datasets, its misclassification rates were much higher compared to all other classifiers considered here. 
{Furthermore, it had the highest misclassification rate in the Herring data set.} TRAD also had relatively higher misclassification rates in many examples (e.g., Synthetic Control Chart, Trace, Toe Segmentation, and Gordon datasets). 
Only in the case of Colon Cancer data, it outperformed others.
Our proposed classifiers had good performance in most of these examples. While the classifiers based on multiple neighbors 
(rMDist, rMDist1 and rMDistC) 
outperformed those based on single neighbors (MDist and MDist1) in
Chowdary, Toe Segmentation, and Colon Cancer data sets, in all other cases, they had comparable performance. In Trace and Herring data sets, these classifiers performed better than TRAD,
TRIPD1 and TRIPD2 classifiers. Table \ref{tab:datasets} clearly shows that the performances of our proposed classifiers, particularly for classifiers based on multiple neighbors, were comparable to nonlinear SVM and random forest. 

\begin{figure}[t]
\centering
\setlength{\tabcolsep}{0pt}
\includegraphics[height=2.50in,width=0.975\linewidth]{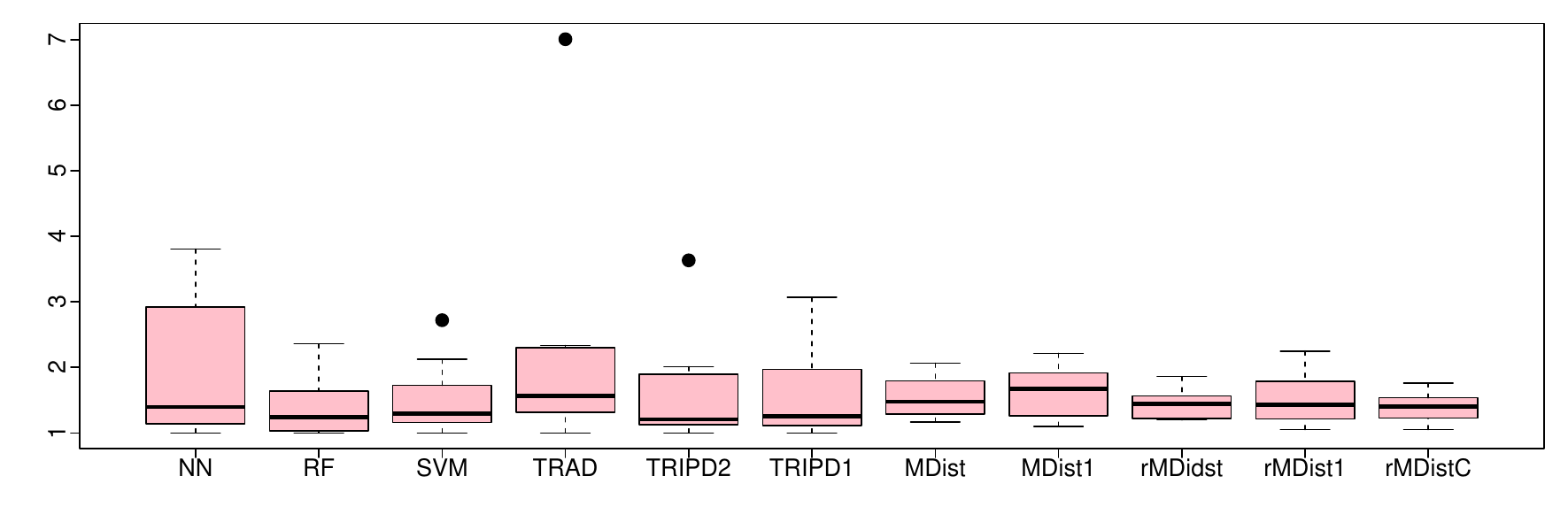}  
\vspace{-0.25in}
\caption{Boxplots showing the robustness of different classifiers in benchmark datasets.\label{fig:boxplot}}
\end{figure}

To compare the overall performances of different classifiers concisely and comprehensively, we used the notion of robustness introduced in \cite{friedman1994flexible}. If there are $T$
classifiers who have misclassification rates $e_1,e_2,\ldots,e_T$ in a particular data set, the robustness of the $t$-th classifier is computed as $R(t)=e_t/e_0$, where $e_0=\min_{1 \le t \le T} e_t$.
So, in an example, the best classifier has $R(t)=1$, while  higher values of $R(t)$
indicate the lack of robustness of the $t$-th classifier. For each
of these benchmark data sets, we computed these ratios  for all
classifiers, and they are graphically represented using
box plots in Figure \ref{fig:boxplot}. This figure clearly shows that the overall performances of all other classifiers were somewhat better than the usual nearest neighbor classifier. It also shows that among our proposed classifiers, those based on multiple neighbors performed better than the corresponding classifiers based on a single nearest neighbor. While the rMDist classifier exhibited better robustness properties than TRIPD2, the rMDist1 classifier
turned out to be more robust than the TRIPD1 classifier.
The rMDistC classifier, which considers both $\ell_1$ and $\ell_2$ distances of the nearest neighbors, also had an excellent overall performance.
If not better, the performances of our proposed classifiers were comparable to the popular classifiers like nonlinear SVM and random forest.

\section{Concluding remarks}

In this article, we have proposed some possible modifications to the nearest neighbor classifier for the classification of high-dimensional data. We have seen that if the location difference among the competing classes gets masked by their scale difference, the usual nearest neighbor classifier performs poorly in high dimensions. The adjustment proposed by \cite{chan2009scale} takes care of this problem, but the resulting classifier fails when the competing classes differ outside their first moments. The MCH classifier overcomes this limitation, and it can discriminate between two high-dimensional distributions differing either in their locations or in their scales. However, this method may not work well in many situations, especially when the class distributions are mixtures of several widely varying sub-classes. 
The proposed classifiers based on minimum distances are helpful in such situations. The MDist1 classifier can even discriminate among competing classes differing outside the first two moments. Instead of considering only one neighbor from each class, sometimes it is helpful to consider the distances of the first $r$ neighbors and perform nearest neighbor classification in that feature space. Analyzing several simulated and benchmark datasets, we have amply demonstrated that if not better, our proposed classifiers yield competitive performance in high dimensions.

In this article, we have used
nearest neighbor classification on the feature space of $\ell_1$ or $\ell_2$ distances. 
Though we have seen some theoretical advantages of using the $\ell_1$ distance, our analysis of benchmark datasets clearly shows that in practice, there is no clear winner. So, a user may wonder which of the two feature spaces to use in a
given problem. One may also like to use $\ell_p$-distances for other choices of $p$ or features based on other generalized distance functions of the form $\varphi_{h,\psi}(\boldsymbol x,\boldsymbol y) = h\big\{\frac{1}{d}\sum_{i=1}^d \psi\big(|x^{(i)} - y^{(i)}|^2\big)\big\}$ where $h: \R_{+}\rightarrow \R_{+}$ and $\psi: \R_{+}\rightarrow \R_{+}$ are continuous, strictly increasing functions with $h(0)=\psi(0) = 0$ as introduced in \cite{sarkar2018gdist}. For suitable choices of $h$ and $\psi$ (e.g., when $h(t) = t$ and $\psi(t)$ has non-constant completely monotone derivatives), it ensures the positivity of the energy distance (as discussed in the paragraph before Theorem 3) in any finite dimension. Moreover, if $\psi$ is bounded, it also makes the resulting classifier robust against outliers. However, the choice of the optimal features is a challenging problem, and it would be helpful if a data-driven method can be developed for this purpose. Throughout this article, for our proposed methods, we have always used the $1$-NN classifier in the feature space. This is mainly for a fair comparison with other competing nearest neighbor methods (e.g., CH, TRAD, TRIPD1, and TRIPD2), where $1$-NN classification is considered. However, in practice, one may use the $k$-NN classifier for other values of $k$ as well. For constructing the rMDistC
classifier, though we have considered the same number of $\ell_1$ and $\ell_2$ distances as features, it is possible to include $r_1$ many $\ell_1$ distances and $r_2$ {many} $\ell_2$ distance in the set of features. 
We avoid choosing different values of $r_1$ and $r_2$ to reduce the computing cost at the cross-validation step. In practice,
distances from all of the first $r$ neighbors may not always be important
for classification. In such cases, a suitable feature selection criterion would be helpful. Instead of feature selection, one can also think about constructing an ensemble classifier \citep[see, e.g.][]{dietterich2000ensemble,zhang2009novel,kiziloz2021classifier} like random forest, where we construct different classifiers based on different sets of features and judiciously aggregate them. These problems can be investigated in a separate work in future.

\vspace{12pt}
%








\bibliographystyle{sa}
\bibliography{sample}

\appendix
\section*{\hfill Appendix \hfill}
\noindent
{\bf Lemma 1:} If  $J$ competing classes satisfy (A1)-(A3), for two independent random vectors $\Xvec \sim F_j$ and $\Yvec \sim F_i$ ($1\le j,i \le J$), $\|\Xvec- \Yvec\|^2/d \stackrel{P}{\rightarrow} \sigma_j^2+\sigma_i^2+\nu_{ji}^2$, where $\nu_{ji}^2=0$ for $j=i$.

\noindent
{\bf Proof:} Note that using Chebyshev's inequality, for any $\epsilon>0$, we get
$$P\left(\Big|\frac{1}{d}\|\Xvec-\Yvec\|^2-E\Big(\frac{1}{d}\|\Xvec-\Yvec\|^2\Big)\Big| \ge \epsilon\right)\leq \frac{1}{\epsilon^2}Var\left(\frac{1}{d}\|\Xvec-\Yvec\|^2\right).$$
Now, $Var\left(\frac{1}{d}\|\Xvec-\Yvec\|^2\right) =\frac{1}{d^2}\left[\disp\sum_{s=1}^{d} Var((X_s-Y_s)^2) + \disp\sum_{s=1}^{d}\sum_{t=1, t\neq s}^{d} Cov\Big((X_s-Y_s)^2,(X_t-Y_t)^2\Big)\right]$
Since the measurement variables from all classes have uniformly bounded fourth moments (see (A1)), we have $\disp\sum_{s=1}^{d} Var((X_s-Y_s)^2)=O(d)$. 
Also, one can show that under assumptions  (A1) and (A2),
$\disp\sum_{s=1}^{d}\sum_{t=1, t\neq s}^{d} Cov\Big((X_s-Y_s)^2,(X_t-Y_t)^2\Big)=o(d^2)$. So, $Var\left(\frac{1}{d}\|\Xvec-\Yvec\|^2\right) \rightarrow 0$ and hence
$\Big|\frac{1}{d}\|\Xvec-\Yvec\|^2-E\Big(\frac{1}{d}\|\Xvec-\Yvec\|^2\Big)\Big| \stackrel{P}{\rightarrow} 0$  as $d \rightarrow \infty$.

Now, $E\Big(\frac{1}{d}\|\Xvec-\Yvec\|^2\Big)=E\Big(\frac{1}{d}\big\|\big(\Xvec-E(\Xvec)\big)-\big(\Yvec-E(\Yvec)\big)+\big(E(\Xvec)-E(\Yvec)\big)\big\|^2\Big)=\frac{1}{d}\mbox{trace}(\sigmat_j) + \frac{1}{d}\mbox{trace}(\sigmat_i) 
+\frac{1}{d}\|\muvec_j-\muvec_i\|^2
\rightarrow \sigma_j^2+\sigma_i^2+\nu_{ji}^2$ as $d \rightarrow \infty$.
So, $\frac{1}{d}\|\Xvec-\Yvec\|^2 \stackrel{P}{\rightarrow}
\sigma_j^2+\sigma_i^2+\nu_{ji}^2$.
Note that if $\Xvec$ and $\Yvec$ follow the same distribution (i.e. $j=i$), we have $\nu_{ji}^2=0.$ \hfill$\Box$

\vspace{0.1in}
\noindent
{\bf Proof of Theorem 1:} (a) From Lemma 1, it is clear that for any test case $\Zvec$ from the $j$-th class ($j=1,2,\ldots,J$), $\frac{1}{d}\|\Zvec-\Xvec_{i\ell}\|^2 \stackrel{P}{\rightarrow} \sigma_j^2+\sigma_i^2+\nu_{ji}^2$ for $i=1,2,\ldots,J$ and $\ell=1,2,\ldots,n_i$. Therefore, for $k<\min\{n_1,n_2,\ldots,n_J\}$,
the $k$-nearest neighbor classifier correctly classifies $\Zvec$ if $2\sigma_j^2<\sigma_j^2+\sigma_i^2+\nu_{ji}^2$ for all $i \neq j$ or equivalently, $\nu_{ji}^2>\sigma_j^2-\sigma_i^2$ for all 
$i\neq j$.
Similarly, for correct classification for a test case from the $i$-th class, we need
$\nu_{ij}^2>\sigma_i^2-\sigma_j^2$ for all 
$j\neq i$. Combining these, we get 
$\nu_{ji}^2>|\sigma_j^2-\sigma_i^2|$ for all $j\neq i$.

If $\nu_{ji}^2<|\sigma_j^2-\sigma_i^2|$ for any pair ($j,i$), we have either $\nu_{ji}^2<\sigma_j^2-\sigma_i^2$ or
$\nu_{ji}^2<\sigma_i^2-\sigma_j^2$. Without loss of generality, let us assume the first one. In that case, for any class-$j$ observation, the distances of its neighbors from the $i$-th class turn out to be smaller than those from $j$-th class with probability tending to $1$ as $d$ increases. So, all observations from the $j$-th class are misclassified with probability tending to $1$.

\vspace{0.05in}
\noindent
(b) From Lemma 1, it is clear that for any test case $\Zvec$ from the $j$-th class, $\frac{1}{d}\rho_j(\Zvec,\Xvec_{j\ell})\stackrel{P}{\rightarrow} \sigma_j^2$ for $\ell=1,2,\ldots,n_j$ whereas for any $i\neq j$,
$\frac{1}{d}\rho_i(\Zvec,\Xvec_{i\ell})\stackrel{P}{\rightarrow} \sigma_j^2+\nu_{ji}^2$ for $\ell=1,2,\ldots,n_i$. So, $\Zvec$ is correctly classified if $\nu_{ji}^2>0$ for all $i \neq j$.
Repeating this argument for $j=1,2,\ldots,J$, we get the result.

\vspace{0.05in}
\noindent
(c) Lemma 1 shows that for any test case $\Zvec$ from the $j$-th class, $\frac{1}{\sqrt{d}}\rho^{\ast}_j(\Zvec,\Xvec_{j\ell})\stackrel{P}{\rightarrow} \sigma_j/\sqrt{2}$ for $\ell=1,2,\ldots,n_j$ whereas for any $i\neq j$,
$\frac{1}{\sqrt{d}}\rho^{\ast}_i(\Zvec,\Xvec_{i\ell})\stackrel{P}{\rightarrow} \sqrt{\sigma_j^2+\sigma_i^2+\nu_{ji}^2}-\sigma_i/\sqrt{2}$ for $\ell=1,2,\ldots,n_i$. So, $\Zvec$ is correctly classified if $\sqrt{\sigma_j^2+\sigma_i^2+\nu_{ji}^2}>(\sigma_j+\sigma_i)/\sqrt{2}$ or $\sigma_j^2+\sigma_i^2+\nu_{ji}^2>(\sigma_j+\sigma_i)^2/{2}$ for all $i \neq j$. Note that $\sigma_j^2+\sigma_i^2+\nu_{ji}^2-(\sigma_j+\sigma_i)^2/{2}= \nu_{ji}^2+(\sigma_j-\sigma_i)^2/2$, which is positive under the given condition. Now, the proof follows by the
repetition of the same argument for $j=1,2,\ldots,J$.  \hfill $\Box$

\vspace{0.1in}
\noindent
{\bf Proof of Theorem 2:} 
%
For the sake of simplicity, let us prove it for $J=2$. For $J>2$, it can be proved similarly. From Lemma 1, for any training sample observation $\Xvec_{1i}$
($i=1,2,\ldots,n_1$) from the first class, as $d$ grows to infinity, we have $$\left( \frac{1}{\sqrt{d}} \min\limits_{1\leq \ell (\neq i)\leq n_1} \|\Xvec_{1i} - \Xvec_{1\ell}\|, \frac{1}{\sqrt{d}} \min\limits_{1\leq \ell \leq n_2} \|\Xvec_{1i} - \Xvec_{2\ell}\|\right) \stackrel{P}{\rightarrow} (\sigma_{1}\sqrt{2}, \sqrt{\sigma_{1}^2+\sigma_{2}^2+\nu^2_{12}})={\bf a}_1, \mbox{ say.}$$ 
Similarly, for a training sample observation $\Xvec_{2i}$
($i=1,2,\ldots,n_2$) from the second class, as $d$ tends to infinity, we have 
$$\left( \frac{1}{\sqrt{d}} \min_{1\leq \ell \leq n_1} \|\Xvec_{2i} - \Xvec_{1\ell}\|, \frac{1}{\sqrt{d}} \min_{1\leq \ell (\neq i) \leq n_2} \|\Xvec_{2i} - \Xvec_{2\ell}\|\right) \stackrel{P}{\rightarrow} ( \sqrt{\sigma_{1}^2+\sigma_{2}^2+\nu^2_{12}}, \sigma_{2}\sqrt{2})={\bf a}_2,  \mbox{ say}.$$
So, if ${\bf a}_1 \neq {\bf a}_2$, the feature vectors obtained from two sets of training sample observations converge to two distinct points ${\bf a}_1$ and ${\bf a}_2$, respectively. 
Now, for any test case $\Zvec$, using Lemma  1, it can be shown that as $d$ grows to infinity,
$( \frac{1}{\sqrt{d}} \min\limits_{1\leq i\leq n_1} \|\Zvec - \Xvec_{1i}\|, \frac{1}{\sqrt{d}} \min\limits_{1\leq i\leq n_2} \|\Zvec - \Xvec_{2i}\|)$ converges in probability to ${\bf a}_1$ and ${\bf a}_2$ for $\Zvec \sim F_1$ and $\Zvec \sim F_2$, respectively.

Therefore, for any $\Zvec\sim F_1$ (respectively, $F_2$), in the $d_1-d_2$ space, while the scaled versions of its distances from the feature vectors from Class-$1$ (respectively, Class-$2$) converge to $0$, those from the feature vectors from Class-$2$ (respectively, Class-$1$)
converge to $\|{\bf a}_1-{\bf a}_2\|$ as $d$ tends to infinity. So, it is correctly classified with probability tending to $1$. Therefore, for  perfect classification by the MDist classifier, we need ${\bf a}_1$
and ${\bf a}_2$ to be distinct, i.e., $2\sigma_1^2$, $2 \sigma_2^2$ and $\sigma_1^2+\sigma_2^2+\nu_{12}^2$ cannot be {all} equal. Note that these three quantities are equal if and only if $\nu_{12}^2=0$ and $\sigma_1^2=\sigma_2^2$, 
{which cannot happen under the assumptions of Theorem 2.} \hfill $\Box$



\vspace{0.1in}
\noindent
{\bf Lemma 2:} Suppose that $\Xvec,\Xvec^{\prime}\sim F_1$ and $\Yvec,\Yvec^{\prime} \sim F_2$ are four independent $d$-dimensional random vectors with finite first moments. Then, we have $$2E\|\Xvec-\Yvec\|_1 -E\|\Xvec-\Xvec^{\prime}\|_1 - E\|\Yvec-\Yvec^{\prime}\|_1 \ge 0$$ where the equality holds if and only if $F_1$ and $F_2$ have identical one-dimensional marginals.

\noindent
{\bf Proof:} First note that 
$$2E\|\Xvec_1-\Yvec_1\|_1 -E\|\Xvec_1-\Xvec_2\|_1 - E\|\Yvec_1-\Yvec_2\|_1
= \sum_{q=1}^{d} \big[2E|X_q-Y_q| -E|X_q-X^{\prime}_q| - E|Y_q-Y^{\prime}_q|\big].$$ 
Now, from \cite{baringhaus2004new}, we get
$$2E|X_q-Y_q| -E|X_q-X^{\prime}_q| - E|Y_q-Y^{\prime}_q| =2\int \limits_{-\infty}^{\infty}\big(F_1^{(q)}(t)-F_2^{(q)}(t)\big)^2 dt,$$ where $F_1^{(q)}$ and $F_2^{(q)}$ are the distribution functions of $X_q$ and $Y_q$, respectively. Clearly, it is non-negative, and it takes the value $0$ if and only if $F_1^{(q)}= F_2^{(q)}$, i.e.,  $X_q$ and $Y_q$ have the same distribution. This shows that $2E\|\Xvec_1-\Yvec_1\|_1 -E\|\Xvec_1-\Xvec_2\|_1 - E\|\Yvec_1-\Yvec_2\|_1\ge 0$, where the equality holds if and only if $X_q$ and $Y_q$ have the same distribution for all $q=1,,2\ldots,d$. \hfill $\Box$

\vspace{0.1in}
\noindent
{\bf Proof of Theorem 3:}
 As in the proof of Theorem 2, here also, for the sake of simplicity, we prove the result for $J=2$. For $J>2$, it can be proved similarly. 

Consider two random vectors $\Xvec \sim F_j$ and $\Yvec \sim F_i$ ($1 \le j,i \le 2$). 
Under (A1$^{\circ}$) and (A2$^{\circ})$, we have
$\big|\frac{1}{d}\|\Xvec-\Yvec\|_1 -E\big(\frac{1}{d}\|\Xvec-\Yvec\|_1\big)\big| \stackrel{P}{\rightarrow}0$ as $d \rightarrow \infty$, and under (A3$^{\circ}$), we have $\lim_{d \rightarrow \infty} E\big(\frac{1}{d}\|\Xvec-\Yvec\|_1\big)=\tau_{ji}$. Lemma 2 shows that $2\tau_{12}-\tau_{11}-\tau_{22} \ge 0$ and under the assumption
${\cal E}_{12}>0$, the equality is ruled out. So, here we have 
$2\tau_{12}-\tau_{11}-\tau_{22} > 0$, which implies that $\tau_{11}, \tau_{12}$ and $\tau_{22}$ cannot be equal.

Now note that for any training sample observation $\Xvec_{1i}$
($i=1,2,\ldots,n_1$) from the first class, as $d$ grows to infinity, $$\left( \frac{1}{{d}} \min\limits_{1\leq \ell (\neq i)\leq n_1} \|\Xvec_{1i} - \Xvec_{1\ell}\|_1, \frac{1}{{d}} \min\limits_{1\leq \ell \leq n_2} \|\Xvec_{1i} - \Xvec_{2\ell}\|_1\right) \stackrel{P}{\rightarrow} (\tau_{11}, \tau_{12})={\bf a}^{\circ}_1, \mbox{ say.}$$ 
Similarly, for a training sample observation $\Xvec_{2i}$
($i=1,2,\ldots,n_2$) from the second class, as $d$ tends to infinity, 
$$\left( \frac{1}{{d}} \min_{1\leq \ell \leq n_1} \|\Xvec_{2i} - \Xvec_{1\ell}\|_1, \frac{1}{{d}} \min_{1\leq \ell (\neq i) \leq n_2} \|\Xvec_{2i} - \Xvec_{2\ell}\|_1\right) \stackrel{P}{\rightarrow} ( \tau_{12}, \tau_{22}),={\bf a}^{\circ}_2,  \mbox{ say.}.$$

Since $\tau_{11}$, $\tau_{12}$ and $\tau_{22}$ are not equal,
we have ${\bf a}^{\circ}_1 \neq {\bf a}^{\circ}_2$.
So, the feature vectors obtained from two sets of training sample observations converge to two distinct points ${\bf a}^{\circ}_1$ and ${\bf a}^{\circ}_2$, respectively. 
For a test case $\Zvec$,  as $d$ grows to infinity,
$( \frac{1}{{d}} \min\limits_{1\leq i\leq n_1} \|\Zvec - \Xvec_{1i}\|_1, \frac{1}{{d}} \min\limits_{1\leq i\leq n_2} \|\Zvec - \Xvec_{2i}\|_1)$ converges in probability to ${\bf a}^{\circ}_1$ and ${\bf a}^{\circ}_2$ for $\Zvec \sim F_1$ and $\Zvec \sim F_2$, respectively.

Therefore, for any $\Zvec\sim F_1$ (respectively, $F_2$), in the $d^{\ast}_1-d^{\ast}_2$ space, while the scaled versions of its distances from the feature vectors from Class-$1$ (respectively, Class-$2$) converge to $0$, those from the feature vectors from Class-$2$ (respectively, Class-$1$)
converge to $\|{\bf a}^{\circ}_1-{\bf a}^{\circ}_2\|$. So, it is correctly classified with probability tending to $1$. \hfill $\Box$




\end{document}